\documentclass[11pt,a4paper]{article}
\usepackage[hyperref]{custom-naacl}
\aclfinalcopy
\usepackage{times}
\usepackage{latexsym}
\usepackage{floatpag}

\usepackage[parfill]{parskip}
\usepackage[utf8]{inputenc} 
\usepackage[T1]{fontenc}    
\usepackage{hyperref}       
\usepackage{url}            
\usepackage{booktabs}       
\usepackage{amsfonts}       
\usepackage{nicefrac}       
\usepackage{microtype}      
\usepackage{graphicx}
\usepackage{natbib}
\usepackage{tabularx}
\usepackage{xtab}

\usepackage{pifont}
\newcommand{\cmark}{\ding{51}}%

\usepackage{amsmath}
\usepackage{amsthm}
\usepackage[toc,page]{appendix}
\usepackage{cleveref}
\usepackage{subfig} 
\usepackage{wrapfig}
\usepackage{cleveref}
\usepackage{comment}

\usepackage{times}
\usepackage{latexsym}

\usepackage{adjustbox}
\usepackage{array}

\def\signed #1{{\leavevmode\unskip\nobreak\hfil\penalty50\hskip2em
  \hbox{}\nobreak\hfil#1%
  \parfillskip=0pt \finalhyphendemerits=0 \endgraf}}

\newsavebox\mybox

\title{The Pile: An 800GB Dataset of Diverse Text for Language Modeling}
\author{
    Leo Gao \And Stella Biderman \And Sid Black \And Laurence Golding
    \AND Travis Hoppe \And Charles Foster \And Jason Phang \And Horace He
    \AND Anish Thite \And Noa Nabeshima \And Shawn Presser \And Connor Leahy
    \AND \textnormal{EleutherAI}\\\url{contact@eleuther.ai}
}
\begin{document}
\maketitle

\begin{abstract}
    Recent work has demonstrated that increased training dataset diversity improves general cross-domain knowledge and downstream generalization capability for large-scale language models. With this in mind, we present \textit{the Pile}: an 825 GiB English text corpus targeted at training large-scale language models. The Pile is constructed from 22 diverse high-quality subsets---both existing and newly constructed---many of which derive from academic or professional sources. Our evaluation of the untuned performance of GPT-2 and GPT-3 on the Pile shows that these models struggle on many of its components, such as academic writing. Conversely, models trained on the Pile improve significantly over both Raw CC and CC-100 on all components of the Pile, while improving performance on downstream evaluations. Through an in-depth exploratory analysis, we document potentially concerning aspects of the data for prospective users. We make publicly available the code used in its construction.\footnote{\url{https://pile.eleuther.ai/}}

\end{abstract}

\section{Introduction}

\begin{figure*}
  \includegraphics[width=\linewidth]{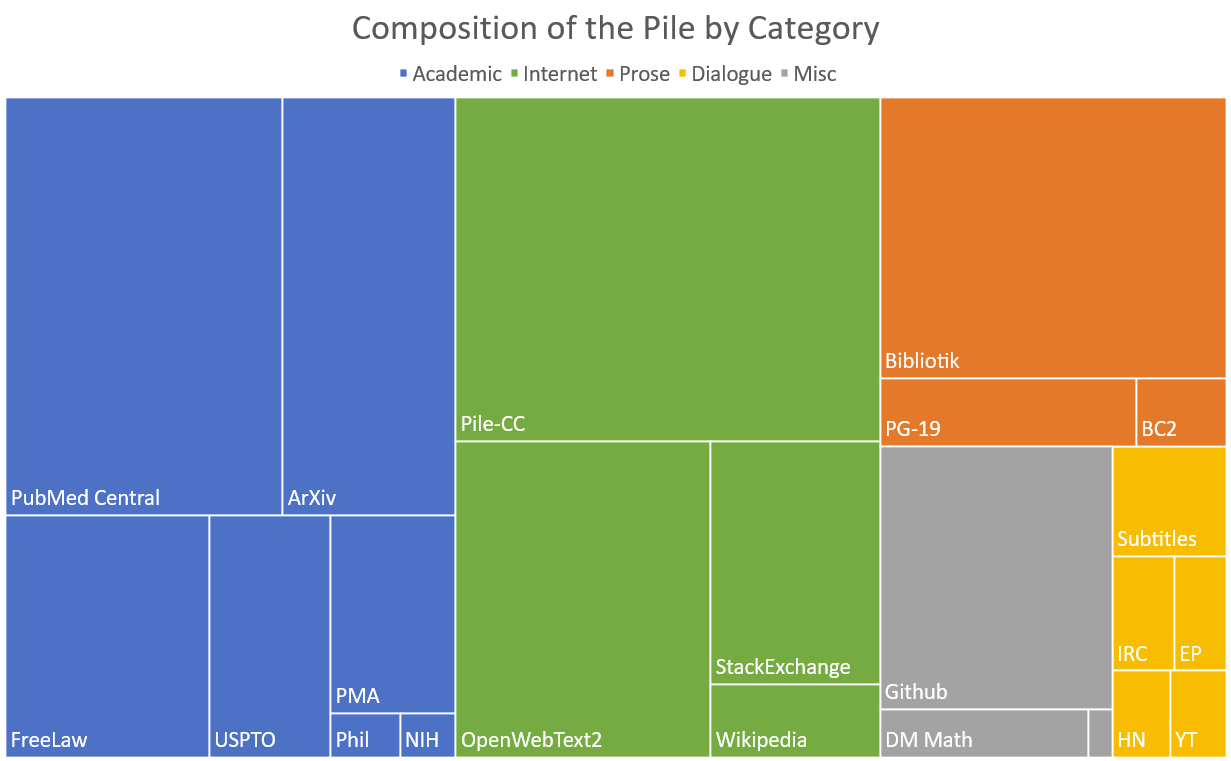}
  \caption{Treemap of Pile components by effective size.}
  \label{fig:proportions}
\end{figure*}

Recent breakthroughs in general-purpose language modeling have demonstrated the effectiveness of training massive models on large text corpora for downstream applications \citep{GPT2,Megatron,T5,TuringNLG,GPT3,GShard}. 
As the field continues to scale up language model training, the demand for high-quality massive text data will continue to grow \citep{scaling-nlms}.

The growing need for data in language modeling has caused most existing large-scale language models to turn to the Common Crawl for most or all of their data \citep{GPT3,T5}. While training on the Common Crawl has been effective, recent work has shown that dataset diversity leads to better downstream generalization capability \citep{TuringNLG}. Additionally, large-scale language models have been shown to effectively acquire knowledge in a novel domain with only relatively small amounts of training data from that domain \citep{TuringNLG,GPT3,extractingtrainingdata}. These results suggest that by mixing together a large number of smaller, high quality, diverse datasets, we can improve the general cross-domain knowledge and downstream generalization capabilities of the model compared to models trained on only a handful of data sources.

To address this need, we introduce the Pile: a $825.18$ GiB English text dataset designed for training large scale language models. The Pile is composed of 22 diverse and high-quality datasets, including both established natural language processing datasets and several newly introduced ones. In addition to its utility in training large language models, the Pile can also serve as a broad-coverage benchmark for cross-domain knowledge and generalization ability of language models.

We introduce new datasets derived from the following sources: PubMed Central, ArXiv, GitHub, the FreeLaw Project, Stack Exchange, the US Patent and Trademark Office, PubMed, Ubuntu IRC, HackerNews, YouTube, PhilPapers, and NIH ExPorter. We also introduce OpenWebText2 and BookCorpus2, which are extensions of the original OpenWebText \citep{OpenWeb} and BookCorpus \citep{BookCorpus,BookCorpusCode} datasets, respectively.

In addition, we incorporate several existing high-quality datasets: Books3 \citep{bibliotik}, Project Gutenberg (PG-19) \citep{PG19}, OpenSubtitles \citep{OpenSubtitles}, English Wikipedia, DM Mathematics \citep{dm-mathematics}, EuroParl \citep{EuroParl}, and the Enron Emails corpus \citep{Enron}. To supplement these, we also introduce a new filtered subset of Common Crawl, Pile-CC, with improved extraction quality.

Through our analyses, we confirm that the Pile is significantly distinct from pure Common Crawl data. Additionally, our evaluations show that the existing GPT-2 and GPT-3 models perform poorly on many components of the Pile, and that models trained on the Pile significantly outperform both raw and filtered Common Crawl models. To complement the performance evaluations, we also perform an exploratory analysis of the text within the Pile to provide a detailed picture of the data. We hope that our extensive documentation of the construction and characteristics of the Pile will help researchers make informed decisions about potential downstream applications.

Finally, we make publicly available the preprocessing code for the constituent datasets of the Pile and the code for constructing alternative versions\footnote{\url{https://github.com/EleutherAI/the-pile}}. In the interest of reproducibility, we also document all processing performed on each dataset (and the Pile as a whole) in as much detail as possible. For further details about the processing of each dataset, see \Cref{sec:data} and Appendix \ref{apdx:data}.

\subsection{Contributions}

The core contributions of this paper are:

\begin{enumerate}
    \item The introduction of a $825.18$ GiB english-language dataset for language modeling combining 22 diverse sources. 
    \item The introduction of $14$ new language modeling datasets, which we expect to be of independent interest to researchers.
    \item Evaluations demonstrating significant improvements across many domains by GPT-2-sized models trained on this new dataset, compared to training on CC-100 and raw Common Crawl.
    \item The investigation and documentation of this dataset, which we hope will better inform researchers about how to use it as well as motivate them to undertake similar investigations of their own data.
\end{enumerate}

\section{The Pile Datasets}\label{sec:data}

\begin{table*}[ht]
    \centering
    \begin{tabular}{l r r r r r}
    \toprule
        \textbf{Component} & \textbf{Raw Size} & \textbf{Weight} & \textbf{Epochs} & \textbf{Effective Size} & \textbf{Mean Document Size} \\
        \midrule
        Pile-CC & 227.12 GiB & 18.11\% & 1.0 & 227.12 GiB & 4.33 KiB \\
        PubMed Central & 90.27 GiB & 14.40\% & 2.0 & 180.55 GiB & 30.55 KiB \\
        Books3\textsuperscript{\textdagger} & 100.96 GiB & 12.07\% & 1.5 & 151.44 GiB & 538.36 KiB \\
        OpenWebText2 & 62.77 GiB & 10.01\% & 2.0 & 125.54 GiB & 3.85 KiB \\
        ArXiv & 56.21 GiB & 8.96\% & 2.0 & 112.42 GiB & 46.61 KiB \\
        Github & 95.16 GiB & 7.59\% & 1.0 & 95.16 GiB & 5.25 KiB \\
        FreeLaw & 51.15 GiB & 6.12\% & 1.5 & 76.73 GiB & 15.06 KiB \\
        Stack Exchange & 32.20 GiB & 5.13\% & 2.0 & 64.39 GiB & 2.16 KiB \\
        USPTO Backgrounds & 22.90 GiB & 3.65\% & 2.0 & 45.81 GiB & 4.08 KiB \\
        PubMed Abstracts & 19.26 GiB & 3.07\% & 2.0 & 38.53 GiB & 1.30 KiB \\
        Gutenberg (PG-19)\textsuperscript{\textdagger} & 10.88 GiB & 2.17\% & 2.5 & 27.19 GiB & 398.73 KiB \\
        OpenSubtitles\textsuperscript{\textdagger} & 12.98 GiB & 1.55\% & 1.5 & 19.47 GiB & 30.48 KiB \\
        Wikipedia (en)\textsuperscript{\textdagger} & 6.38 GiB & 1.53\% & 3.0 & 19.13 GiB & 1.11 KiB \\
        DM Mathematics\textsuperscript{\textdagger} & 7.75 GiB & 1.24\% & 2.0 & 15.49 GiB & 8.00 KiB \\
        Ubuntu IRC & 5.52 GiB & 0.88\% & 2.0 & 11.03 GiB & 545.48 KiB \\
        BookCorpus2 & 6.30 GiB & 0.75\% & 1.5 & 9.45 GiB & 369.87 KiB \\
        EuroParl\textsuperscript{\textdagger}& 4.59 GiB & 0.73\% & 2.0 & 9.17 GiB & 68.87 KiB \\
        HackerNews & 3.90 GiB & 0.62\% & 2.0 & 7.80 GiB & 4.92 KiB \\
        YoutubeSubtitles & 3.73 GiB & 0.60\% & 2.0 & 7.47 GiB & 22.55 KiB \\
        PhilPapers & 2.38 GiB & 0.38\% & 2.0 & 4.76 GiB & 73.37 KiB \\
        NIH ExPorter & 1.89 GiB & 0.30\% & 2.0 & 3.79 GiB & 2.11 KiB \\
        Enron Emails\textsuperscript{\textdagger}& 0.88 GiB & 0.14\% & 2.0 & 1.76 GiB & 1.78 KiB \\
        \midrule
        \textbf{The Pile} & \textbf{825.18 GiB} &  &  & \textbf{1254.20 GiB} & \textbf{5.91 KiB} \\
        \bottomrule

    \end{tabular}
\caption{Overview of datasets in the Pile before creating the held out sets. Raw Size is the size before any up- or down-sampling. Weight is the percentage of bytes in the final dataset occupied by each dataset. Epochs is the number of passes over each constituent dataset during a full epoch over the Pile. Effective Size is the approximate number of bytes in the Pile occupied by each dataset. Datasets marked with a \textdagger\ are used with minimal preprocessing from prior work.} 
\label{table:pile_overview}
\end{table*}

The Pile is composed of 22 constituent sub-datasets, as shown in \Cref{table:pile_overview}. 
Following \citet{GPT3}, we increase the weights of higher quality components, with certain high-quality datasets such as Wikipedia being seen up to 3 times (``epochs'') for each full epoch over the Pile.
Detailed information about the construction of each dataset is available in Appendix \ref{apdx:data}.

\subsection{Pile-CC}

Common Crawl is a collection of website crawls from 2008 onwards, including raw web pages, metadata and text extractions. Due to the raw nature of the dataset, Common Crawl has the advantage of including text from diverse domains, but at the cost of varying quality data. Due to this, use of Common Crawl typically necessitates well-designed extraction and filtering. Our Common Crawl-based dataset, Pile-CC, uses jusText \citep{justext} on Web Archive files (raw HTTP responses including page HTML) for extraction, which yields higher quality output than directly using the WET files (extracted plaintext). 

\subsection{PubMed Central}\label{subsection:pmc}

PubMed Central (PMC) is a subset of the PubMed online repository for biomedical articles run by the United States of America's National Center for Biotechnology Information (NCBI), providing open, full-text access to nearly five million publications. Most publications indexed by PMC are recent, and their inclusion is mandated for all NIH funded research starting from 2008 by the NIH Public Access Policy. We included PMC in the hopes that it will benefit potential downstream applications to the medical domain.

\subsection{Books3}

Books3 is a dataset of books derived from a copy of the contents of the Bibliotik private tracker made available by Shawn Presser \citep{bibliotik}. Bibliotik consists of a mix of fiction and nonfiction books and is almost an order of magnitude larger than our next largest book dataset (BookCorpus2).We included Bibliotik because books are invaluable for long-range context modeling research and coherent storytelling.

\subsection{OpenWebText2}

OpenWebText2 (OWT2) is a generalized web scrape dataset inspired by WebText \citep{GPT2} and OpenWebTextCorpus \citep{OpenWeb}. Similar to the original WebText, we use net upvotes on Reddit submissions as a proxy for outgoing link quality. OpenWebText2 includes more recent content from Reddit submissions up until 2020, content from multiple languages, document metadata, multiple dataset versions, and open source replication code. We included OWT2 as a high quality general purpose dataset.

\subsection{ArXiv}

ArXiv is a preprint server for research papers that has operated since 1991. As shown in \cref{fig:arXiv}, arXiv papers are predominantly in the fields of Math, Computer Science, and Physics. We included arXiv in the hopes that it will be a source of high quality text and math knowledge, and benefit potential downstream applications to research in these areas. ArXiv papers are written in LaTeX, a common typesetting language for mathematics, computer science, physics, and some adjacent fields. Training a language model to be able to generate papers written in LaTeX could be a huge boon to the research community.

\subsection{GitHub}

GitHub is a large corpus of open-source code repositories. Motivated by the ability of GPT-3 \citep{GPT3} to generate plausible code completions despite its training data not containing any explicitly gathered code datasets, we included GitHub in the hopes that it would enable better downstream performance on code-related tasks.

\subsection{FreeLaw}

The Free Law Project is a US-registered non-profit that provides access to and analytical tools for academic studies in the legal realm. CourtListener,\footnote{\url{https://www.courtlistener.com/}} part of the Free Law Project, provides bulk downloads for millions of legal opinions from federal and state courts. While the full dataset provides multiple modalities of legal proceedings, including dockets, bibliographic information on judges, and other metadata, we focused specifically on court opinions due to an abundance of full-text entries. This data is entirely within the public domain.

\subsection{Stack Exchange}

The Stack Exchange Data Dump\footnote{\url{https://archive.org/details/stackexchange}} contains an anonymized set of all user-contributed content on the Stack Exchange network, a popular collection of websites centered around user-contributed questions and answers. It is one of the largest publicly available repositories of question-answer pairs, and covers a wide range of subjects---from programming, to gardening, to Buddhism. We included Stack Exchange in the hopes that it will improve the question answering capabilities of downstream models on diverse domains.

\subsection{USPTO Backgrounds}

USPTO Backgrounds is a dataset of background sections from patents granted by the United States Patent and Trademark Office, derived from its published bulk archives\footnote{\url{https://bulkdata.uspto.gov/}}. A typical patent background lays out the general context of the invention, gives an overview of the technical field, and sets up the framing of the problem space. We included USPTO Backgrounds because it contains a large volume of technical writing on applied subjects, aimed at a non-technical audience.

\subsection{Wikipedia (English)}

Wikipedia is a standard source of high-quality text for language modeling. In addition to being a source of high quality, clean English text, it is also valuable as it is written in expository prose, and spans many domains.

\subsection{PubMed Abstracts}

PubMed Abstracts consists of the abstracts from 30 million publications in PubMed, the online repository for biomedical articles run by the National Library of Medicine. While the PMC (see Section \ref{subsection:pmc}) provides full-text access, the subset of coverage is significantly limited and biased towards recent publications. PubMed also incorporates MEDLINE, which expands the coverage of biomedical abstracts from 1946 to present day.

\subsection{Project Gutenberg}

Project Gutenberg is a dataset of  classic Western literature. The specific Project Gutenberg derived dataset we used, PG-19, consists of Project Gutenberg books from before 1919 \citep{PG19}, which represent distinct styles from the more modern Books3 and BookCorpus. Additionally, the PG-19 dataset is already being used for long-distance context modeling.

\subsection{OpenSubtitles}

The OpenSubtitles dataset is an English language dataset of subtitles from movies and television shows gathered by \citet{OpenSubtitles}. Subtitles provide an important source of natural dialog, as well as an understanding of fictional formats other than prose, which may prove useful for creative writing generation tasks such as screenwriting, speechwriting, and interactive storytelling.

\subsection{DeepMind Mathematics}

The DeepMind Mathematics dataset consists of a collection of mathematical problems from topics such as algebra, arithmetic, calculus, number theory, and probability, formatted as natural language prompts \citep{dm-mathematics}. One major weakness of large language models has been performance on mathematical tasks \citep{GPT3}, which may be due in part to a lack of math problems in the training set. By explicitly including a dataset of mathematical problems, we hope to improve the mathematical ability of language models trained on the Pile.

\subsection{BookCorpus2}

BookCorpus2 is an expanded version of the original BookCorpus \citep{BookCorpus}, a widely used language modeling corpus consisting of books written by ``as of yet unpublished authors.'' BookCorpus is therefore unlikely to have significant overlap with Project Gutenberg and Books3, which consist of published books. BookCorpus is also commonly used as dataset for training language models \citep{GPT, BERT, RoBERTa}.

\subsection{Ubuntu IRC}

The Ubuntu IRC dataset is derived from the publicly available chatlogs\footnote{\url{https://irclogs.ubuntu.com/}} of all Ubuntu-related channels on the Freenode IRC chat server. Chatlog data provides an opportunity to model real-time human interactions, which feature a level of spontaneity not typically found in other modes of social media. 

\subsection{EuroParl}

EuroParl \citep{EuroParl} is a multilingual parallel corpus originally introduced for machine translation but which has also seen use in several other fields of NLP \citep{EuroParl-example1,EuroParl-example2,EuroParl-example3}. We use the most current version at time of writing, which consists of the proceedings of the European Parliament in 21 European languages from 1996 until 2012.

\subsection{YouTube Subtitles}

The YouTube Subtitles dataset is a parallel corpus of text gathered from human generated closed-captions on YouTube. In addition to providing multilingual data, Youtube Subtitles is also a source of educational content, popular culture, and natural dialog.

\subsection{PhilPapers}

The PhilPapers\footnote{\url{https://philpapers.org/}} dataset consists of open-access philosophy publications from an international database maintained by the Center for Digital Philosophy at the University of Western Ontario. We included PhilPapers because it spans a wide body of abstract, conceptual discourse, and its articles contain high quality academic writing.

\subsection{NIH Grant Abstracts: ExPORTER}

The NIH Grant abstracts provides a bulk-data repository for awarded applications through the {ExPORTER}\footnote{\url{https://exporter.nih.gov/}} service covering the fiscal years 1985-present. We included the dataset because it contains examples of high-quality scientific writing.

\subsection{Hacker News}
Hacker News\footnote{\url{https://news.ycombinator.com}} is a link aggregator operated by Y Combinator, a startup incubator and investment fund. Users submit articles defined as ``anything that gratifies one's intellectual curiosity,'' but submitted articles tend to focus on topics in computer science and entrepreneurship. Users can comment on submitted stories, resulting in comment trees discussing and critiquing submitted stories. We scrape, parse, and include these comment trees since we believe they provide high quality dialogue and debate on niche topics. 

\subsection{Enron Emails}

The Enron Emails dataset \citep{Enron} is a valuable corpus commonly used for research about the usage patterns of email. We included Enron Emails to aid in understanding the modality of email communications, which is typically not found in any of our other datasets.

\section{Benchmarking Language Models with the Pile}

While the Pile was conceived as a training dataset for large-scale language models, its coverage of multiple disparate domains makes it also suitable as an evaluation dataset.
In this section, we describe how the Pile can be used as a broad-coverage dataset for benchmarking language models.

\subsection{Benchmarking Guidelines}

The Pile is provided as train, validation, and testing splits. The validation and testing components each contain $0.1\%$ of the data, sampled uniformly at random. While this is a far smaller percentage than most datasets, the sheer size of the dataset results in over 1 GiB of validation and testing data each. We highlight that while we have made efforts to deduplicate documents within the Pile (See: Section~\ref{apdx:deduplication}), it is still possible that some documents are duplicated across the train/validation/test splits.

Our preferred metric is bits per UTF-8 encoded byte (\textsc{bpb}). Bits per byte is preferred over bits per character or perplexity when using Pile as a metric due to its invariance to different tokenization schemes and the ambiguity of measuring characters in Unicode. To compute bits per byte from a given negative log likelihood loss $\ell$, we compute $\textsc{bpb} = (L_T/L_B) \log_2(e^\ell) = (L_T/L_B) \ell / \ln(2)$, where $L_T$ is the length of the dataset in tokens and $L_B$ is the length of the dataset in UTF-8 encoded bytes. We find that $L_T/L_B$ is $0.29335$ GPT-2-tokens/byte across the Pile; dataset-specific values of $L_T/L_B$ can be found in Table \ref{tbl:bpb_conversion}.

\subsection{Test Perplexity with GPT-2 and GPT-3}

We compute the test perplexity of the constituent datasets of the Pile using GPT-2 \citep{GPT2} and GPT-3 \citep{GPT3}, shown in Figure~\ref{table:pile_perplexity_utf8}. We use all available versions of GPT-2, and all four versions of GPT-3 available via the OpenAI API. Because of the cost associated with using the OpenAI API, we evaluate on one-tenth of the respective test sets for most of the constituent datasets. We report the perplexity converted to bits per UTF-8 encoded byte (\textsc{bpb}). Importantly, we compute perplexity by evaluating each document independently within each dataset, as opposed to concatenating all documents as is common practice for computing perplexity on large corpora.

Full details of the perplexity computation can be found in Appendix~\ref{apdx:perplexity}.

\begin{figure}[ht]
  \includegraphics[width=\linewidth]{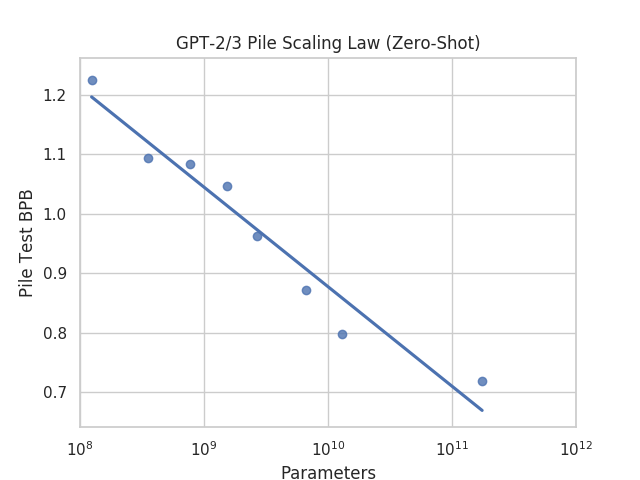}
  \caption{Scaling law for performance of GPT-2/3 models. `Zero-shot' refers to the fact that none of the models have been fine-tuned on data from the Pile.}
  \label{fig:scaling_law_gpt3}
\end{figure}

Unsurprisingly, larger language models generally attain lower perplexity compared to smaller models. Recent work has shown an increased focus on the empirical scaling laws of language models \citep{scaling-nlms, scaling-autoregressive}. As such, we investigate the scaling law for the GPT-2 and GPT-3 families of models on perplexity evaluation on the Pile. The scaling law relation for the GPT-3 family of models is shown in Figure~\ref{fig:scaling_law_gpt3}.\footnote{While the sizes of GPT-3 models on the OpenAI API have not been publicized, we assume here that \texttt{ada}, \texttt{babbage}, \texttt{curie} and \texttt{davinci} models correspond to 2.7B, 6.7B, 13B and 175B parameter models respectively.} The line of best fit shown in the figure has a coefficient of -0.1674 and an intercept of 2.5516.

Interestingly, while GPT-2 and GPT-3 were not trained on the Pile, there still appears to be a clear scaling law without diminishing returns. We hypothesize that this is due to the inherent generalization capability of these models. We leave a more rigorous analysis of zero-shot scaling laws to future work. 

\begin{table*}[ht]
\resizebox{\textwidth}{!}{\small%
    \centering
    \begin{tabular}{l rrrr rrrr}
    \toprule
        \textbf{Component} 
        & \multicolumn{4}{c}{GPT-2}
        & \multicolumn{4}{c}{GPT-3}
        \\ \cmidrule(lr){2-5} \cmidrule(lr){6-9}
        & \multicolumn{1}{c}{small}
        & \multicolumn{1}{c}{medium}
        & \multicolumn{1}{c}{large}
        & \multicolumn{1}{c}{xl}
        & \multicolumn{1}{c}{ada}
        & \multicolumn{1}{c}{babbage}
        & \multicolumn{1}{c}{curie}
        & \multicolumn{1}{c}{davinci}
        \\
        \midrule
    Pile-CC & 1.0878 & 0.9992 & 0.9582 & 0.9355 & 0.9212 & 0.8483 & 0.7849 & \textbf{0.7070} \\
    PubMed Central & 1.0759 & 0.9788 & 0.9334 & 0.9044 & 0.8633 & 0.7792 & 0.7150 & \textbf{0.6544} \\
    Books3 & 1.1959 & 1.1063 & 1.0588 & 1.0287 & 0.9778 & 0.9005 & 0.8284 & \textbf{0.7052} \\
    OpenWebText2 & 1.1111 & 1.0073 & 0.9539 & 0.9171 & 0.8727 & 0.7921 & 0.7199 & \textbf{0.6242} \\
    ArXiv & 1.3548 & 1.2305 & 1.1778 & 1.1381 & 1.0304 & 0.9259 & 0.8453 & \textbf{0.7702} \\
    Github & 1.7912 & 1.3180 & 1.7909 & 1.6486 & 0.8761 & 0.7335 & 0.6415 & \textbf{0.5635} \\
    FreeLaw & 1.0512 & 0.9321 & 0.9017 & 0.8747 & 0.8226 & 0.7381 & 0.6667 & \textbf{0.6006} \\
    Stack Exchange & 1.2981 & 1.1075 & 1.0806 & 1.0504 & 1.0096 & 0.8839 & 0.8004 & \textbf{0.7321} \\
    USPTO Backgrounds & 0.8288 & 0.7564 & 0.7202 & 0.6969 & 0.6799 & 0.6230 & 0.5752 & \textbf{0.5280} \\
    PubMed Abstracts & 0.9524 & 0.8579 & 0.8108 & 0.7810 & 0.8130 & 0.7382 & 0.6773 & \textbf{0.6201} \\
    Gutenberg (PG-19) & 1.2655 & 1.1140 & 1.0820 & 1.0829 & 0.9776 & 0.8749 & 0.7930 & \textbf{0.7115} \\
    OpenSubtitles & 1.2465 & 1.1657 & 1.1324 & 1.1129 & 1.1116 & 1.0488 & 0.9875 & \textbf{0.9130} \\
    Wikipedia (en) & 1.1285 & 1.0213 & 0.9795 & 0.9655 & 0.8757 & 0.7863 & 0.7047 & \textbf{0.5953} \\
    DM Mathematics & 2.6911 & 2.5448 & 2.4833 & 2.4377 & 2.3249 & 2.2015 & 2.1067 & \textbf{2.0228} \\
    Ubuntu IRC & 1.8466 & 1.7187 & 1.6427 & 1.6024 & 1.3139 & 1.1968 & 1.0995 & \textbf{0.9915} \\
    BookCorpus2 & 1.1295 & 1.0498 & 1.0061 & 0.9783 & 0.9754 & 0.9041 & 0.8435 & \textbf{0.7788} \\
    EuroParl & 2.3177 & 2.0204 & 1.8770 & 1.7650 & 1.0475 & 0.9363 & 0.8415 & \textbf{0.7519} \\
    HackerNews & 1.4433 & 1.2794 & 1.3143 & 1.3361 & 1.1736 & 1.0875 & 1.0175 & \textbf{0.9457} \\
    YoutubeSubtitles & 2.0387 & 1.8412 & 1.7355 & 1.6694 & 1.3407 & 1.1876 & 1.0639 & \textbf{0.9469} \\
    PhilPapers & 1.3203 & 1.2163 & 1.1688 & 1.1327 & 1.0362 & 0.9530 & 0.8802 & \textbf{0.8059} \\
    NIH ExPorter & 0.9099 & 0.8323 & 0.7946 & 0.7694 & 0.7974 & 0.7326 & 0.6784 & \textbf{0.6239} \\
    Enron Emails & 1.5888 & 1.4119 & 1.4535 & 1.4222 & 1.2634 & 1.1685 & 1.0990 & \textbf{1.0201} \\
    \midrule
    The Pile & 1.2253 & 1.0928 & 1.0828 & 1.0468 & 0.9631 & 0.8718 & 0.7980 & \textbf{0.7177} \\
        \bottomrule
    \end{tabular}
}
\caption{Test perplexity of the Pile using GPT-2 and GPT-3, converted to bits per UTF-8 encoded byte (\textsc{bpb}). Evaluation is performed on one-tenth of the test data of the Pile, on a per-document basis. \textbf{Bold} indicates the best-performing model in each row.}
\label{table:pile_perplexity_utf8}
\end{table*}

\subsection{Relative Componentwise GPT-3 Pile Performance}\label{sec:entropy_adjusted}

Determining which components GPT-3 underperforms on provides information about which Pile components are most dissimilar to the distribution of text (web pages and books) that GPT-3 was trained on. These components would thus make especially good candidates for supplementing GPT-3 training data. These results are also valuable for determining which types of datasets to emphasize for future iterations of the Pile.

Due to the difference in entropy of different datasets, directly comparing perplexity of GPT-3 on different Pile components is not an accurate indication of relative performance. Ideally we would train a GPT-3 model from scratch on the Pile and compare the difference in loss per dataset with that of the original GPT-3. Because of resource constraints, we instead use a GPT-2 model trained from scratch on the Pile (see \Cref{sec:ablation}) to construct a proxy measure. To construct our proxy, we first measure the improvement from the GPT-2-Pile model to GPT-3 on each component. Then, we normalize our results by setting the change on OpenWebText2 to be zero. This computation is shown in the equation below:

\begin{align*}
    \Delta_{\mathrm{set}} &= \left(L^{\mathrm{GPT3}}_{\mathrm{set}} - L^{\mathrm{GPT3}}_{\mathrm{owt2}}\right)\\
    & - \left(L^{\mathrm{GPT2Pile}}_{\mathrm{set}} - L^{\mathrm{GPT2Pile}}_{\mathrm{owt2}}\right)
\end{align*}

Since GPT2-Pile was trained on both OWT2 and the dataset we are evaluating, we expect the second term in $\Delta_{set}$ to reflect the difference in the intrinsic difficulty of the two datasets. Thus the total value of $\Delta_{set}$ reflects how much harder the dataset we are evaluating was for GPT-3 than OWT2, minus the relative difficulty of the two tasks. As GPT-3 was trained on data very similar to OWT2, this gives us a proxy for how much better GPT-3 would do if it were trained on the Pile.

\begin{figure*}[ht]
  \includegraphics[width=\linewidth]{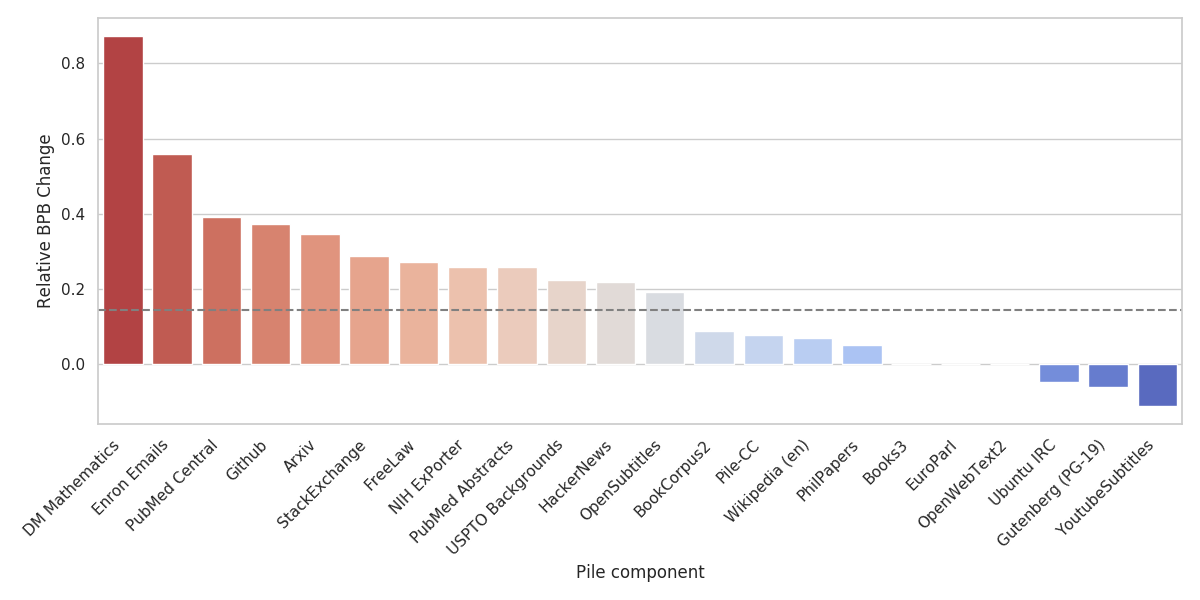}
  \caption{Change in \textsc{bpb} from GPT-2 trained on Pile to GPT-3 zero-shot, relative to OpenWebText2 \textsc{bpb} change. Dotted line indicates overall Pile change. Lower indicates better relative performance by GPT-3.}
  \label{fig:rel_bpb_gpt3} 
\end{figure*}

The results are shown in Figure~\ref{fig:rel_bpb_gpt3}. As a sanity check, we observe that datasets that are contained in, or are extremely similar to, GPT-3's training set (Books3, Wikipedia (en), Pile-CC and Project Gutenberg) score close to zero on our metric.

GPT-3 appears to perform poorly on datasets pertaining to research or academic writing like PubMed Central, PubMed Abstracts, and ArXiv; domain-specific datasets like FreeLaw, HackerNews, and USPTO Backgrounds; and on datasets containing predominantly text distinct from natural language, like GitHub and DM Mathematics. In addition, the majority of datasets see less of an improvement than OpenWebText2. As such, we expect a GPT-3 sized model trained on Pile to perform significantly better on research related tasks, software tasks, and symbol manipulation tasks than the base model. Additionally, this experiment provides evidence that the majority of Pile components are not redundant with the predominantly web-based GPT-3 training data.

We note that this metric is only a proxy for similarity, and that it could be confounded by dataset specific scaling effects. Although our results largely accord with expectations, there are some puzzling results, like the datasets on which GPT-3 outperformed GPT-2 Pile. We hypothesize that GPT-3 learns to be so good at these datasets that training on them explicitly does not notably benefit the model's performance. We leave a more rigorous analysis of these effects for future work.

\newcolumntype{R}[2]{%
    >{\adjustbox{angle=#1,lap=\width-(#2)}\bgroup}%
    l%
    <{\egroup}%
}
\newcommand*\rot{\multicolumn{1}{R{45}{1em}}}

\section{Evaluation}\label{sec:ablation}

\begin{table*}[t]
    \centering
    \begin{tabular}{l r r r r r r r}
    \toprule
         & Dataset Size & Pile (val) & Pile (test) & WikiText & LAMBADA & LAMBADA  \\
         & & \multicolumn{1}{c}{(\textsc{bpb})}& \multicolumn{1}{c}{(\textsc{bpb})}& \multicolumn{1}{c}{(\textsc{ppl})}& \multicolumn{1}{c}{(\textsc{ppl})}& \multicolumn{1}{c}{(\textsc{acc})} \\
        \midrule
        The Pile & 825 GiB & \textbf{0.9281} & \textbf{0.9433} & \textbf{5.59} & 12.78 & \textbf{50.1} \\
        CC-100 (en) & 300 GiB & 1.3143 & 1.3293 & 8.27 & \textbf{11.78} & 49.7 \\
        Raw CC & 45927 GiB\rlap{$^\dagger$} & 1.1180 & 1.1275 & 11.75 & 19.84 & 43.8 \\

        \bottomrule

    \end{tabular}
\caption{Size-controlled evaluation results. Each dataset is deduplicated against all evaluation metrics and subsampled to approximately 40GB to control for the effects of dataset size. For LAMBADA, we use the variant of the data introduced in \citet{GPT2} and only evaluate the perplexity on the final token rather than the final word. For WikiText, we report the perplexity per GPT-2 token. \textdagger\ indicates that the size is an estimate.} 
\label{table:ablation_overview}
\end{table*}

\begin{figure*}[ht]
  \includegraphics[width=\linewidth]{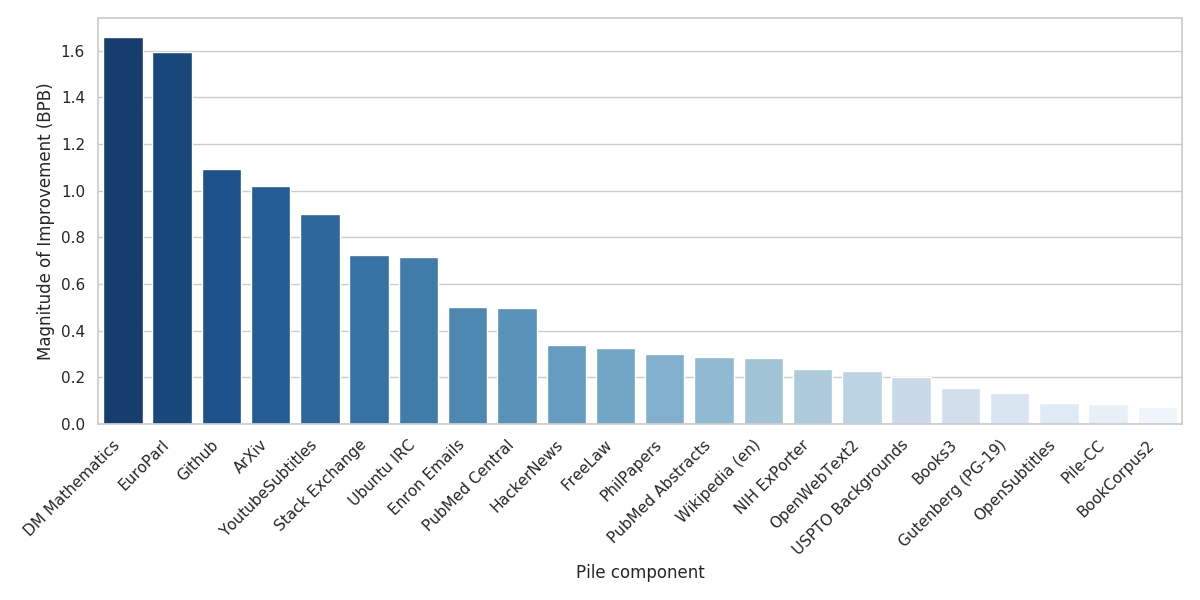}
  \caption{Magnitude of \textsc{bpb} improvement of Pile model over CC-100 model on each test set.}
  \label{fig:pile_improvement_cc100}
\end{figure*}

To confirm the effectiveness of the Pile for improving language modeling quality, we train architecturally-identical 1.3 billion parameter models based on those in \citet{GPT3} on different datasets and evaluate on the WikiText and LAMBADA tasks as benchmarks of language modeling ability. We also report results on the Pile as a measure of more cross-domain generalization.

\subsection{Methodology}

To ensure a fair comparison across datasets of different sizes, we decontaminate any instances of the evaluation sets using the same 13-gram overlap filtering as in \citet{GPT3} and downsample to 40GB to control for dataset size. As we control for dataset size, we emphasize that our evaluation is generous to CC-100 (en), which is about 1/3 the size of the Pile in reality.  

We compare the following datasets: the Pile, the English component of the CC-100 dataset\footnote{The data was obtained from \url{http://data.statmt.org/cc-100/}.} \citep{wenzek2019ccnet,conneau2019xlmr}, and a sample of raw CC WET files filtered for English-only. 

\subsection{Results}

On traditional language modeling benchmarks, the Pile improves significantly on WikiText and shows negligible changes in LAMBADA. However, models trained on Pile improve significantly over both Raw CC and CC-100 on all components of the Pile, as shown in Table \ref{table:ablation_overview_test}. This indicates that models trained on the Pile have greater cross-domain generalization capabilities without compromising performance on traditional benchmarks.

The magnitude of improvement over CC-100 per set is shown in Figure \ref{fig:pile_improvement_cc100}. Unsurprisingly, there is almost no improvement on Pile-CC. However, the model trained on the Pile performs significantly better than either of the other models on academic datasets such as ArXiv, Pubmed Central, FreeLaw, and PhilPapers. It also improves significantly on programming-related datasets like Github and StackExchange, on EuroParl, due to the lack of multilingual text in either other dataset, and on DM Mathematics, indicating a significant improvement in mathematical ability.

Surprisingly, raw Common Crawl performs better on the Pile \textsc{bpb} than CC-100, despite losing by a significant margin on LAMBADA and WikiText. We hypothesize that this is due to the perplexity based filtering used in CC-100, where a language model is trained on Wikipedia and all data with a perplexity too high or too low is discarded. This effectively discards any data too similar to or too different from Wikipedia, which severely limits the diversity of the collected data. This result suggests that future work using Common Crawl should take caution with filtering to preserve its diversity.

\begin{table}[ht]
\resizebox{0.5\textwidth}{!}{\small%
    \centering
    \begin{tabular}{l rrr}
    \toprule
        \textbf{Dataset} 
        & \multicolumn{1}{c}{The Pile}
        & \multicolumn{1}{c}{CC-100 (en)}
        & \multicolumn{1}{c}{Raw CC (en)}
        \\
        \midrule
    Pile-CC & \textbf{0.9989} & 1.0873 & 1.0287 \\
    PubMed Central & \textbf{0.6332} & 1.1311 & 0.9120 \\
    Books3 & \textbf{1.0734} & 1.2264 & 1.1366 \\
    OpenWebText2 & \textbf{0.9938} & 1.2222 & 1.0732 \\
    ArXiv & \textbf{0.7945} & 1.8159 & 1.2642 \\
    Github & \textbf{0.5597} & 1.6509 & 0.9301 \\
    FreeLaw & \textbf{0.6978} & 1.0221 & 0.9468 \\
    Stack Exchange & \textbf{0.8152} & 1.5414 & 1.1292 \\
    USPTO Backgrounds & \textbf{0.6731} & 0.8772 & 0.8455 \\
    PubMed Abstracts & \textbf{0.7313} & 1.0193 & 0.9718 \\
    Gutenberg (PG-19) & \textbf{1.1426} & 1.2780 & 1.2235 \\
    OpenSubtitles & \textbf{1.0909} & 1.1827 & 1.2139 \\
    Wikipedia (en) & \textbf{0.8961} & 1.1807 & 1.0252 \\
    DM Mathematics & \textbf{1.5206} & 3.1774 & 2.6229 \\
    Ubuntu IRC & \textbf{1.4085} & 2.1243 & 1.5691 \\
    BookCorpus2 & \textbf{1.0613} & 1.1346 & 1.0914 \\
    EuroParl & \textbf{1.1202} & 2.7141 & 1.4917 \\
    HackerNews & \textbf{1.0968} & 1.4352 & 1.2305 \\
    YoutubeSubtitles & \textbf{1.4269} & 2.3287 & 1.5607 \\
    PhilPapers & \textbf{1.1256} & 1.4269 & 1.2090 \\
    NIH ExPorter & \textbf{0.7347} & 0.9713 & 0.9225 \\
    Enron Emails & \textbf{0.8301} & 1.3300 & 1.0483 \\

        \bottomrule
    \end{tabular}
}
\caption{Breakdown of \textsc{bpb} on Pile heldout test set. Columns indicate the dataset each model is trained on; rows indicate the evaluation dataset. \textbf{Bold} indicates the best performing model in each row.}
 \label{table:ablation_overview_test}
\end{table}

\section{Structural Statistics}

In this section, we cover the Structural Statistics of the dataset, which provide more coarse-grained and statistical information about the Pile. In Section~\ref{sec:doc}, we provide a closer investigation and documentation of the textual content within the Pile datasets.

\subsection{Document Lengths and Tokenization}

Each dataset consists of a large number of documents. We analyze the distribution of document lengths, as well as the number of bytes-per-token using the GPT-2 tokenizer in order to put our ablations in context. 

\begin{figure}[ht]
  \includegraphics[width=\linewidth]{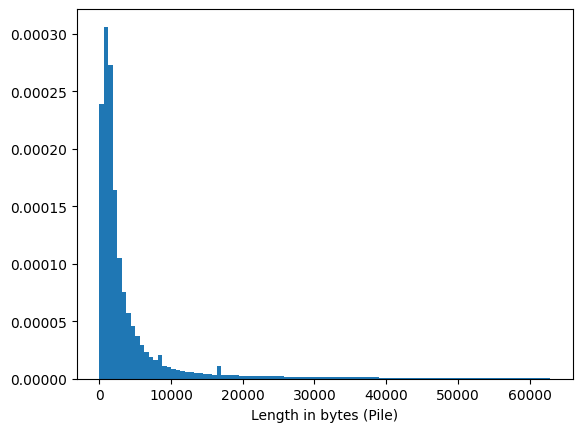}
  \caption{Distribution of document lengths in Pile. The highest 1 percentile of document length are considered to be outliers and excluded from this plot.}
  \label{fig:len_bytes}
\end{figure}

While the majority of documents in the Pile are short, there is a long tail of very long documents (Figure \ref{fig:len_bytes}). 

Since the GPT-2 BPE tokenizer is trained on WebText, the mean bytes per token is also a very rough indicator of how syntactically different each Pile component is from WebText. For instance, datasets like NIH ExPorter, OpenWebText2 and Books3 consist largely of ordinary text in a similar distribution to WebText, which is reflected in a greater number of bytes per token. On the other hand, many of the sets with the lowest bytes per token are those which consist in large part of non-text content (Github, ArXiv, Stack Exchange, and DM Mathematics) or languages other than English (EuroParl).

\begin{figure}[ht]
  \includegraphics[width=\linewidth]{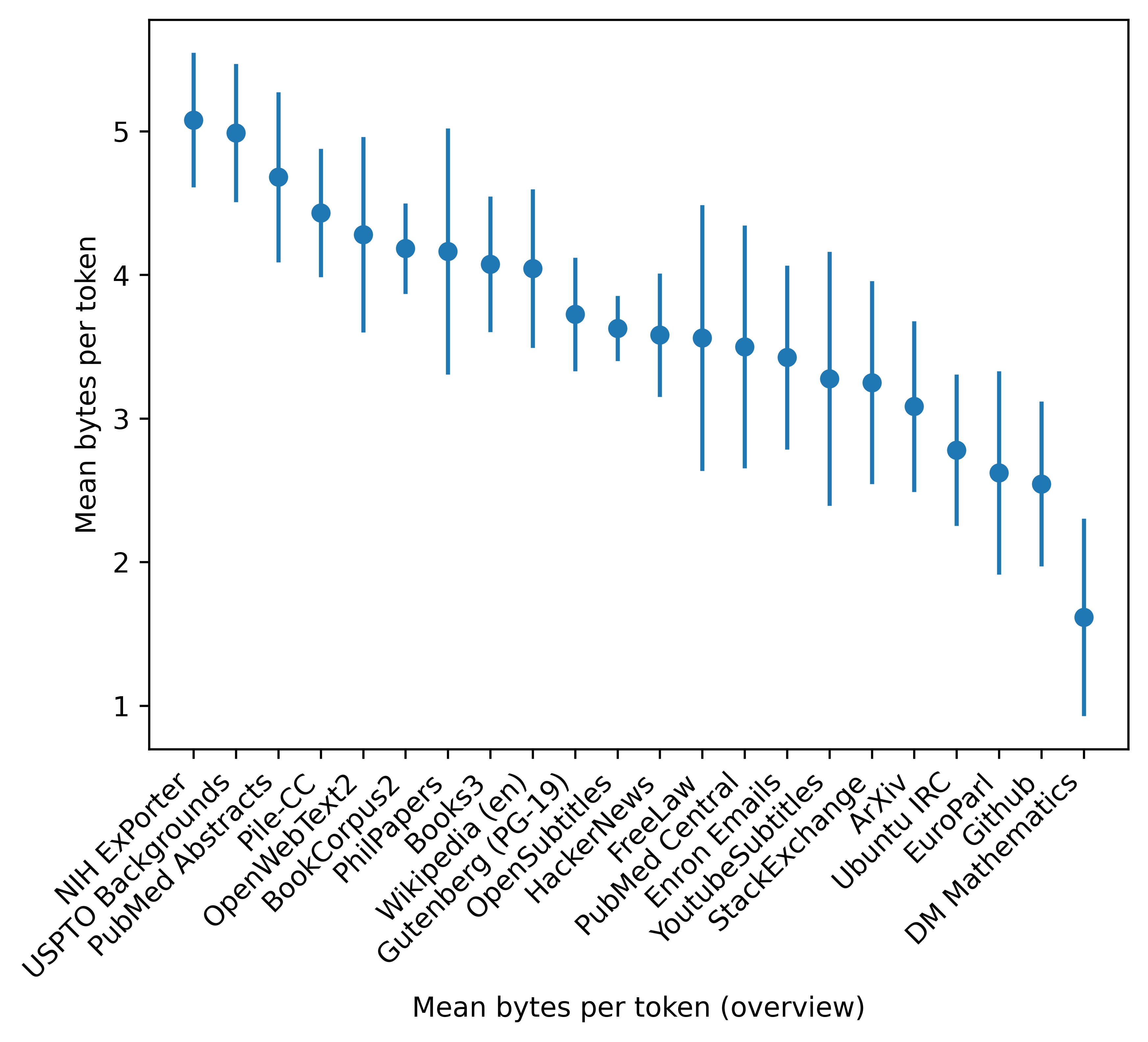}
  \caption{Mean bytes per GPT-2-token for each dataset in the Pile. Error bars indicate standard deviation.}
  \label{fig:bytes_per_token}
\end{figure}

\subsection{Language and Dialects}

While only 13\% of the world's population speaks English, the vast majority of NLP research is done on English. For the Pile, we took a similar approach to the dataset used by \citet{GPT3} and focused predominantly on English, while also not explicitly filtering out other languages when collecting our own data. When evaluating a multilingual dataset, our main criteria for inclusion was whether the English component of the dataset merited inclusion alone. We plan to create a fully multilingual expansion of the Pile as future work.

Using fasttext \cite{fast-text}, we determine that the Pile is 97.4\% English. We note that due to issues with language identification, particularly with rare languages \citet{caswell2020language}, this methodology provides only a rough estimate for English content and no reliable conclusions for low-resource languages can be drawn.

\section{Investigating and Documenting the Datasets}\label{sec:doc}

As the scale of machine learning research has grown, scrutiny has been placed on the ever larger datasets that models are trained on \citep{pyrrhic-cv,best-practice2}

While this issue has been raised within AI ethics and bias research \citep{nlp-impact,disability-bias,nlp-bias-survey}, it has not been a focal point of concern within the language modeling community. Despite the proliferation of work exploring and documenting issues with datasets \citep{datasheets,datastatement,archives}, no dataset intended to train massive language models has been seriously documented by its creators\footnote{\citet{GPT3} discusses ethical issues surrounding their \textit{model}, but do not discuss those surrounding the training dataset itself.}. Therefore, our analyses serve two goals: to address ethical concerns about the Pile, and to promote and normalize the practice of engaging with the AI ethics literature.

Natural language processing technologies are widely applicable and can be used in extremely different contexts. What is and is not appropriate data to train on can therefore vary wildly with the application context. In our view, the best approach is to \textit{document} rather than \textit{eliminate} potentially concerning aspects of datasets\footnote{That said, we did exclude several datasets, see Appendix \ref{apdx:excluded} for details.}, particularly since the purpose of the Pile is to train general-purpose language models. The primary goal of our documentation, therefore, is to empower NLP researchers to make informed decisions.

\subsection{Documenting Methods}

To document the Pile, we chose to implement two frameworks that have been proposed by methodologists and ethics researchers. The first, the datasheets methodology \citep{datasheets}, is a general purpose methodology that is recommended by several methodologists \citep{best-practice1,best-practice2} and appears to be used more frequently by practitioners than alternatives \citep{datasheet1,datasheet2,datasheet3}. The second, the data statements methodology \citep{datastatement}, was proposed specifically for natural language processing and has been well received by the NLP community. Our datasheet and data statement will be featured in the GitHub repository where the code for the Pile is stored and will also be available as separate documents on arXiv \citep{pile-datasheet,pile-datastatement}.

In addition to the datasheet and data statement, there is additional information that may be helpful to people training language models that these documents do not cover. In the rest of this section we investigate and document in greater detail some of this additional contextual information.

\subsection{Topical Distribution}

In order to better understand the specific subject matter covered by the Pile, we performed a topic modeling analysis on its components. Using Gensim \citep{gensim}, we trained 16-topic Latent Dirichlet Allocation \citep{lda} models on each component of the validation set of the Pile concurrently, in an online fashion \citep{onlinelda}. We filtered the Pile for English only for this analysis. Afterwards, we computed the perplexity of the Common Crawl-derived (Pile-CC) topic model on the document sets of the other components. In this way, we provide a rough measure of the degree to which parts of the Pile contain topics not well covered within Common Crawl.

\begin{figure*}[ht]
  \includegraphics[width=\linewidth]{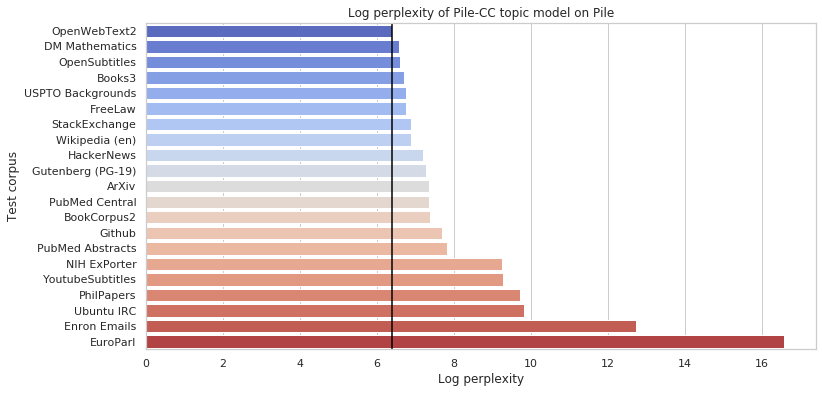}
  \caption{Log perplexity of 16-topic LDA trained on Pile-CC, on other Pile components. Dotted line indicates log perplexity of the topic model on OpenWebText2. Higher indicates a larger topical divergence from Pile-CC.}
  \label{fig:topic_model_perplexities}
\end{figure*}

In Figure \ref{fig:topic_model_perplexities}, these cross-component perplexities are shown, with a vertical line indicating the perplexity of the Pile-CC topic model evaluated on the documents of OpenWebText2. This component was chosen as a baseline of comparison for similar reasons as in the previous evaluation: it is derived in a similar manner (filtered crawls of the open web) as the Common Crawl, and thus is expected to contain a similar distribution of topics. Although Pile-CC is somewhat diverse in its content, several of the Pile's other components deviate from it strongly in their topical focus, as evidenced by higher perplexity on Github, PhilPapers, and EuroParl.

We also documented the topical clusters inferred from our LDA models for each component, which we provide in Appendix \ref{apdx:data}. As expected, though the larger CC-derived component itself represents a diversity of content---including politics, education, sports and entertainment---the content clusters it misses become apparent when compared qualitatively to other components of the Pile. Notably, the data modes covering programming, logic, physics, and legal knowledge appear largely absent.

\subsection{Pejorative Content}
Due to the wide diversity in origins, it is possible for the Pile to contain pejorative, sexually explicit, or otherwise objectionable content. As this content may not be desirable for some use cases, we break down profanity on a per-dataset level.

We used the \texttt{profanity-checker} Python package \cite{profanity-checker}. This package includes a ``toxicity model'' trained on multiple profanity lists as well as the Wikidetox Toxic Comment Dataset \citep{wiki-detox} and classifies a given string as being profane or not profane.

We considered only the English sentences in each dataset using the same language classifier from Section 3.7. We did this since \texttt{profanity-checker} is built for English and other languages may improperly impact the results. For instance, the German nominative/accusative feminine/plural definite article "die" is flagged as being profane regardless of context. We split each sentence into words and computed the percentage of words that are flagged as profane for each component of the Pile. We emphasize that this methodology is only a proxy for profanity, given the complexity of determining whether a given word or phrase is profane in context.

\begin{figure}[ht]
  \includegraphics[width=\linewidth]{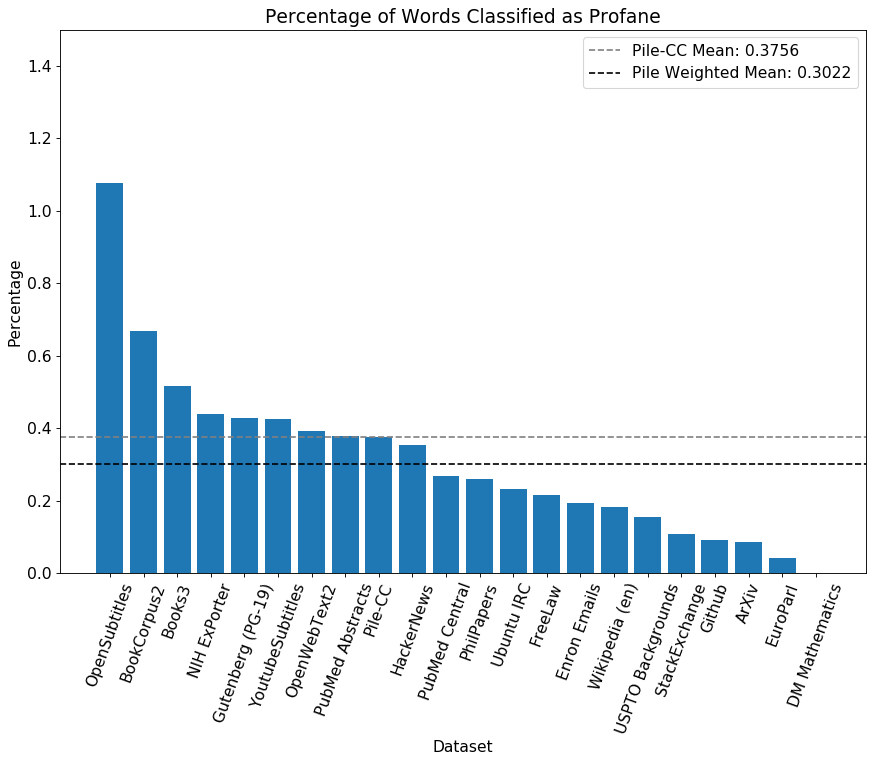}
  \caption{Percentage of words classified as profane in the Pile. The percentage of the CC component and the weighted mean of the Pile as a whole are shown as horizontal lines.}
  \label{fig:profane_words}
\end{figure}

As shown in Figure \ref{fig:profane_words}, the Pile as a whole appears less profane than Pile-CC. Further, the majority of Pile components appear less profane than Pile-CC as well.

We also broke each dataset down on a sentence level, to allow \texttt{profanity-checker} to check entire sentences. Splitting datasets by sentence allows for additional context to be considered when determining whether content is pejorative. Our results are shown in Figure  \ref{fig:profane_sent}.

\subsection{Bias and Sentiment Co-occurrence}
As language models may pick up unexpected biases from the training data, we performed a preliminary analysis of the different components that make up the Pile. Because models with different characteristics may be trained on the Pile, we aimed to document the biases of the data and not a specific model. We primarily focus on co-occurrence tests, where we analyzed what words occur in the same sentence as other specific words. Using this information, we can estimate what words strongly bias towards a category word, as well as calculate the general sentiment of surrounding words.

We focused our analysis on gender, religion, and race. Our goal is to provide users of this dataset with preliminary guidance on how the different components are biased so that they can make decisions on which components to train on. 

All tables and figures in this section can be found in the Appendix.

\subsubsection{Gender}
We computed gender associations by computing co-occurrences for binary pronouns. For each word, we computed the difference in the rate it co-occurs with "he" and "she"\footnote{We chose to only study male and female pronouns as a simplifying assumption. Studying ``they'' would require us to isolate its usage as a singular noun. } and weighed it by the square root of its frequency. We report the top 15 most biased adjectives or adverbs~\citep{nltk} for each in Table~\ref{table:gender_cooccurrence}. We see that words like ``military'', ``criminal'', and ``offensive'' strongly bias towards men, while ``little'', ``married'', ``sexual'', and ``happy'' bias towards women.

In addition, we computed the average sentiment~\citep{sentiment} of words co-occurring with the gendered pronouns across each dataset in Figure~\ref{fig:gender_sentiment}. Generally, we find no significant sentiment bias towards men or women. This, of course, does not mean that the dataset is free of gender bias (as our co-occurrence tests show).

\subsubsection{Religion}
We computed a similar co-occurrence analysis for religion, which can be found in Table \ref{table:religion_cooccurrence}. Like gender, we find that these co-occurrences reflect how these terms are used in pockets of online discourse. For example, ``radical'' co-occurs with ``muslim'' at a high rate, while ``rational'' often co-occurs with ``atheist''. This analysis also demonstrates some of the limitations of a purely co-occurrence based analysis. For example, ``religious'' often co-occurs with ``atheist'', which likely reflects the type of conversations in which the word ``atheist'' is likely to occur as opposed to a descriptor of ``atheist''.

In addition, we computed the average sentiment of co-occurrences across each of the constituent datasets in Figure \ref{fig:religion_sentiment}. Over the entire dataset, we find that ``Buddhist'' has the highest sentiment, followed by ``Hindu'',  ``Christian'', ``Atheist'', and ``Muslim''. Notably, ``Jew'' is the lowest, perhaps reflecting its historical use as a pejorative.

\subsubsection{Race}
Finally, we ran the same analysis for racial groups. Here, as identifiers like ``black'' or ``white'' often do not indicate race, we instead compute co-occurences with phrases like ``black man'' or ``white woman''.

We show the top 15 most biased words for each demographic in Table \ref{table:race_cooccurrence}. Once again, we found that the co-occurrences reflect the context in which these terms are used. For example, the 4 most biased words for ``black'' are ``unarmed'', ``civil'', ``criminal'', and ``scary''.

Similar to above, we compute the average sentiment of co-occurring words. We report the average sentiment numbers in Table \ref{table:race_sentiment}. We find that ``hispanic/latino'' narrowly edges out ``asian'' for the highest sentiment, followed by ``white''. On the other hand, ``black'' had the lowest sentiment, at -0.15.

We note that for all demographics, the average sentiment is negative. We hypothesize that this is due to the specific context for which the phrases we use to compute co-occurrences appear. For example, it is often quite common for news articles to describe suspects as an ``asian man''.

\subsection{Author Consent and Public Data}

Another issue with the use of texts in natural language processing research is consent. Although one is typically not legally obligated to receive the permission of an author to train a NLP algorithm on their work\footnote{Laws vary by country. For a discussion of US law, see \Cref{subsec:law}}, many consider doing so a moral obligation or a good measure to guard against misuse \citep{ethics-consent,pyrrhic-cv}. On the other hand, there is significant disagreement surrounding the ethics of repurposing data protected by \textit{terms of service} in research contexts \cite{beyondthebelmont, norobots}, particularly given the power asymmetries inherent in digital platforms, which often close off independent researchers from investigating public data while simultaneously compelling users to consent to its private use \cite{overcomingtos}.

While much of the Pile's data comes from sources that have expressly consented to its wider dissemination and use in research, researchers often fail to clearly document where their data came from and under what terms its use was consented to. In light of this, we felt it appropriate to release the Pile with transparency around how the authors of its data have indicated that that data can be used.

To provide needed nuance to our discussion of consent, we identified three tiers of availability for public use. \textbf{Public data} is data which is freely and readily available on the internet. This primarily excludes data which is pay-walled (regardless of how easy that paywall is to bypass) and data which cannot be easily obtained but can be obtained, e.g. through a torrent or on the dark web. \textbf{Terms of Service (ToS) compliant data} is data which is obtained and used in a fashion that is known to be consistent with the terms of service of the data host. \textbf{Data with authorial consent} is data for which the original authors of the work consented to the use of their data, or where a reasonable person could not assume that their data would not be used for purposes such as research. ToS compliant data and authorial consented data differ in two main ways: It is important to keep in mind that people typically do not read Terms of Service, and additionally that being ToS-compliant does not entail authorial consent. We adopted a strict model of consent, where ambiguous or unknown consent is treated as non-consensual.

\Cref{table:consent} summarizes our understanding of the status of each of the datasets within the Pile. Datasets marked with a \cmark are compliant in the relevant respects, though a couple datasets are worth remarking on in particular. Book3 and OpenSubtitles are being used in a fashion that is consistent with the terms of service \textit{of the data host}. However, this is somewhat misleading in that the data host is not authorized to post the data online by the parties that own it. The Enron Emails dataset was not collected with the permission of the authors, but was collected by the U.S. government as part of a criminal investigation. While the people whose emails are in the Enron dataset are aware of this fact, they were not given the ability to consent to its inclusion in any way.

There are five datasets included in the Pile that were not collected and distributed in a ToS compliant fashion and for which the authors had no ability to consent to their data being used. Each of these datasets are widely used, both in the NLP literature and the world at large. With the exception of the YouTube Subtitles dataset, each of these datasets were published by researchers and are passed around freely on the internet. The YouTube Subtitles dataset was created by us for this project, using a very popular unofficial API that is both widely used and easily obtainable on Pip, Conda, and GitHub, among other places. Given the processing applied and the difficulty of identifying particular files in the Pile, we feel that our use of these datasets does not constitute significantly increased harm beyond that which has already been done by the widespread publication of these datasets.

\begin{table}
    \centering
    \begin{tabularx}{\columnwidth}{l|c c c}
        \hline
        \textbf{Component} & \textbf{Public} & \textbf{ToS} & \textbf{Author} \\
        \midrule
        Pile-CC & \cmark & \cmark & \\
        PMC & \cmark & \cmark & \cmark\\
        Books3 & \cmark & & \\
        OWT2 & \cmark & & \\
        ArXiv & \cmark & \cmark & \cmark\\
        Github & \cmark & \cmark & \\
        FreeLaw & \cmark & \cmark & \cmark\\
        Stack Exchange & \cmark & \cmark & \cmark\\
        USPTO & \cmark & \cmark & \cmark\\
        PubMed & \cmark & \cmark & \cmark\\
        PG-19 & \cmark & \cmark & \\
        OpenSubtitles & \cmark & & \\
        Wikipedia & \cmark & \cmark & \cmark\\
        DM Math & \cmark & \cmark & \cmark\\
        Ubuntu IRC & \cmark & \cmark & \cmark\\
        BookCorpus2 & \cmark & & \\
        EuroParl & \cmark & \cmark & \cmark\\
        HackerNews & \cmark & \cmark & \\
        YTSubtitles & \cmark & & \\
        PhilPapers & \cmark & \cmark & \cmark\\
        NIH & \cmark & \cmark & \cmark\\
        Enron Emails & \cmark & \cmark & \\
        \hline
    \end{tabularx}
    \caption{Types of consent for each dataset}
    \label{table:consent}
\end{table}

\section{Implications and Broader Impacts}

The Pile represents yet another stepping stone along the path of scaling models and datasets to ever larger sizes and capabilities. There are many serious concerns about how the emergence of progressively stronger AI systems will influence the wider world \citep{brundage_malicious_2018,amodei_concrete_2016,bostrom2014ethics,bostrom2014superintelligence,critch2020ai}, and we believe that they merit serious thought. In this section we discuss the legal ramifications of the Pile, and then consider the impact of the Pile to AI alignment from two angles: accelerating AI timelines and the dangers posed by unaligned language models.

\subsection{Legality of Content}\label{subsec:law}

While the machine learning community has begun to discuss the issue of the legality of training models on copyright data, there is little acknowledgment of the fact that the processing and distribution of data owned by others may also be a violation of copyright law. As a step in that direction, we discuss the reasons we believe that our use of copyright data is in compliance with US copyright law.\footnote{This discussion does not, and is not intended to, constitute legal advice; rather, it is a general discussion of law. Only your attorney can provide assurances that the information contained herein is applicable or appropriate to a particular situation. If in doubt, it is always advisable to speak to an intellectual property attorney.}

Under \citet{presumptive-fair-use} (and affirmed in subsequent rulings such as \citet{affirm-presumption,guild-v-google}), non-commercial, not-for-profit use of copyright media is preemptively fair use. Additionally, our use is \textit{transformative}, in the sense that the original form of the data is ineffective for our purposes and our form of the data is ineffective for the purposes of the original documents. Although we use the full text of copyright works, this is not necessarily disqualifying when the full work is necessary \citep{full-work}. In our case, the long-term dependencies in natural language require that the full text be used in order to produce the best results \citep{transformer-xl,PG19,scaling-autoregressive,liu2018generating}.

Copyright law varies by country, and there may be additional restrictions on some of these works in particular jurisdictions. To enable easier compliance with local laws, the Pile reproduction code is available and can be used to exclude certain components of the Pile which are inappropriate for the user. Unfortunately, we do not have the metadata necessary to determine exactly which texts are copyrighted, and so this can only be undertaken at the component level. Thus, this should be be taken to be a heuristic rather than a precise determination.

\subsection{Acceleration of AI Timelines}

There is serious concern that AI systems may soon be meaningfully more capable than humans in all relevant economic tasks \citep{grace_when_2018,IEM}. Relatedly, there are serious unresolved questions surrounding how to properly align such powerful AI systems with human interests \citep{bostrom2014ethics,russell2019human,bostrom2014superintelligence,amodei_concrete_2016} and generally avoid morally catastrophic outcomes \citep{sotala_superintelligence_2017,monster}. As such, it has been argued that accelerating the development of such powerful AI systems may be undesirable before these concerns have been more adequately addressed \citep{bostrom2014superintelligence}. 

There are several pragmatic responses to this view: 
\begin{enumerate}
\item Due to human competition, curiosity, and cultural diversity, halting technological development is incredibly difficult, if not impossible. \citep{russell2019human} \citep{critch2020ai}
\item AI development is experimental in nature: The alignment problem can only be solved through development, testing and (hopefully non-existential) failure.
\item High powered language models, along with their more general successors, must be capable of viewing morally problematic content without adopting it in their output. We elaborate on this in the following section.
\end{enumerate}

With this in mind, we accept the reality that the Pile could potentially accelerate AI timelines. However, we hope our efforts to establish best practices, such as thoroughly documenting the contents of our data, will help encourage diligence for downstream researchers on alignment problems.

\subsection{Negative LM Output}

There has been much discussion about the possible negative effects of powerful language models in the world \citep{GPT3,brundage_malicious_2018}. Some of these possible problems, such as the ability to mass produce low quality content for the purpose of Search Engine Optimization, are inherent problems to the way online content is distributed, and cannot be stopped by those developing language models alone. Directly solving these problems would require sweeping changes to the architecture of the Internet, such as vastly expanded Public Key Infrastructure and distributed authentication of identity \citep{practical_cryptography}. 

Another concern is that training such models on huge datasets will almost inevitably require them to have undesirable content in their training sets, such as that promoting hateful stereotypes \citep{christian2020alignment}. Having models output undesirable content is, by definition, undesirable, but we believe that attacking this problem from the training set side is unproductive and ultimately leads us away from optimal solutions. 
If a person reads a racist piece of content, they do not then immediately adopt its racist views---they may be capable of doing so, but can decide not to.
This capacity to understand undesirable content and then decide to ignore it is an essential future research direction. Not only would this allow models to use ``dirtier" data with less concern, but also to use their gained knowledge to better understand what not to do. We recognize that, despite recent progress in human-guided learning \citep{learningsum}, the technology is not yet at this stage, and have thus made a number of editorial decisions as described in this paper. However, this approach seems essential to the future of these models and AI more broadly, and more research is needed.

\section{Related Work}

Self-supervised training of natural language processing models on large, unlabeled text corpora, has seen widespread adoption in the field. Word representation models such as GloVe \citep{pennington2014glove} and word2vec \citep{mikolov2013word2vec} were trained on datasets such as Wikipedia, Gigaword \citep{graff2003english}, or a non-public Google News corpus. More recently, language models \citep{GPT,GPT2,GPT3,TuringNLG,Megatron} and masked language models \citep{BERT,RoBERTa,T5} have been trained on datasets such as Wikipedia, BookCorpus \citep{BookCorpus}, RealNews \citep{zellers2019neuralfakenews}, CC-Stories \citep{trinh2018commonsense}, and other Internet scrape-derived datasets discussed below. Other datasets such as WikiText \citep{wikitext} have also been used in similar self-supervised training. 

As data requirements for language modeling have grown, the field has turned towards Internet scrapes for large-scale datasets \citep{OpenWeb}, with Common Crawl being particularly prevalent. Works such as \citet{GPT3,wenzek2019ccnet,OSCAR,T5} have relied on Common Crawl to build training datasets for large-scale models. However, these works often highlight the difficulty of cleaning and filtering the Common Crawl data, and often highlight the resulting data quality as a determining factor of model capability.

It has also been increasingly common practice to combine multiple datasets when training language models. For instance, GPT \citep{GPT} was trained on Wikipedia and BookCorpus, whereas GPT-3 \citep{GPT3} was trained on Wikipedia, two fiction datasets, and two web-scraped datasets. The Pile continues the trend of combining large-scale web-scrapes with smaller, higher-quality datasets that capture knowledge we believe would be most beneficial to training language models.

The two most comparable publicly available datasets to the Pile are CC-100 \citep{wenzek2019ccnet} and C4/mC4 \citep{T5}. C4 is comparably-sized to the Pile, while mC4 and CC-100 are larger, multilingual datasets.
However, C4/mC4 require immense computational resources to preprocess the data, with its maintainers even recommending the use of a distributed cloud service,\footnote{\url{https://www.tensorflow.org/datasets/catalog/c4}} setting a high bar of entry to using these datasets. 
CC-100 is directly downloadable and pre-cleaned; however, its English portion is much smaller than the Pile.
Importantly, these three datasets are all derived entirely from Common Crawl---as discussed above, the current best practice in training large-scale language models involve using both large web scrapes and more targeted, higher-quality datasets, which the Pile directly addresses.

\section{Acknowledgments}

The authors would like to thank TensorFlow Research Cloud for providing the computational resources for the evaluation and OpenAI for providing access and credits for the OpenAI API for GPT-3 evaluation.

We would also like to thank Farrukh Raman, JR Smith, and Michelle Schmitz for reviewing the manuscript.

\bibliography{the-pile-citations}
\bibliographystyle{acl_natbib}

\clearpage

\begin{appendices}

\section{Contributions}\label{apdx:contributions}

All authors contributed to the design of the research project and the writing of the paper. Additionally, authors contributed as follows:

\textbf{Leo Gao} led the project, implemented the main Pile codebase, contributed to the model training code, performed the evaluations and the language analysis, interpreted the perplexity analysis results, implemented the processing to create the final data, and processed Pile-CC, PubMed Central, ArXiv, and Ubuntu IRC.\\
\textbf{Stella Biderman} led the data analysis, the broader impact analysis, and the data documentation, and coordinated the project. She also wrote the analysis of structural statistics, authorial consent, and copyright law.\\
\textbf{Sid Black} implemented the model training and evaluation code and processed YouTube Subtitles, Stack Exchange, and GitHub. \\
\textbf{Laurence Golding} implemented deduplication, performed the n-gram analysis, and processed OpenWebText2. \\
\textbf{Travis Hoppe} processed FreeLaw, Pubmed Abstracts, ExPorter, and PhilPapers. \\
\textbf{Charles Foster} performed the topic modeling analysis, contributed to the discussion of authorial consent, and processed USPTO Backgrounds. \\
\textbf{Jason Phang} implemented and performed the GPT-2/3 perplexity analysis and advised the project. \\
\textbf{Horace He} performed the bias and sentiment analysis. \\
\textbf{Anish Thite} implemented and performed the profanity analysis and processed Hacker News. \\
\textbf{Noa Nabeshima} processed GitHub. \\
\textbf{Shawn Presser} processed BookCorpus2.\\
\textbf{Connor Leahy} wrote the alignment implication analysis and the model training code.

\section{Excluded Datasets}\label{apdx:excluded}

In the course of building the Pile, we considered including and ultimately decided to not use several datasets. We excluded several datasets on the grounds that they were too small to be worth spending time on or because the English component of the data did not merit inclusion on its own. However we also decided to exclude several data sets for other reasons, which we document here for transparency:

\begin{enumerate}
    \item \textbf{US Congressional Record.} The official record of the United States Congress ($1800$ -- today) records important points of debate at the highest levels of American government. It reflects the opinions and biases of the political class over the past 200 years, including segregationism and xenophobia. In particular, we found a large quantity of extremely racist content that we did not feel appropriate for a dataset intended for general-purpose language modeling.
    \item \textbf{Fanfiction.} Hundreds of GiB of fanfiction has been written and put online, primarily on the websites \url{www.fanfiction.net} and \url{www.https://archiveofourown.org/}. This represents a significant untapped resource for language modeling as it is almost exclusively short-form fiction, a writing style that is not represented in most language modeling datasets. We ultimately decided to exclude fanfiction on logistical grounds: we found other sources of data that were easier to obtain.
    \item \textbf{Literotica.} Literotica is a website where users can upload short-form erotic fiction. We had originally planned on including it in the Pile and even went as far as scraping and processing it. However we decided to not include it for several reasons. Firstly, once we decided to exclude fanfiction, Literotica represented our sole source of short-form fiction, which would likely lead to undesirable biases in the trained model. Secondly, Literotica would require significantly more investigation, assessment, and care than we spent on the other datasets. Thirdly, Literotica contains a significant amount of stereotyping, including racial fetishes. While Literotica is likely usable for some tasks, we are not comfortable including it in the Pile.
\end{enumerate}

\section{Dataset Details}\label{apdx:data}

This section contains additional information about each dataset listed in \Cref{sec:data}, including how it was obtained, how it was processed, and any other details relevant for replication. The intent of this section is to provide as much detail as possible, so that Pile can be replicated in the future if necessary, and so that any future processing of these and similar datasets can use or improve on our methods. As such, all code created for processing has been made publicly available under permissive open source licenses and is referenced in footnotes where applicable.

\subsection{Pile-CC}
We extract Common Crawl using jusText \cite{justext}. Our filtering implementation uses a classifier trained against the OpenWebText2 dataset. We process only a small fraction of the available Common Crawl data; we break the list of urls to individual WARC files from 2013 to 2020 into 3679 chunks and process 22 random chunks.

\subsubsection{WARC vs WET}

CommonCrawl data is available in two main formats: Web ARChive (WARC) files, which contain a full record of the crawl as well as the raw HTML of the webpage, and WET files, which contain pre-extracted versions of the contents of the WARC files. The WET files have poor quality, often containing large amounts of boilerplate text like menus and page footers, but due to the lower bandwidth and computation requirements necessary to use WET files, prior work based on CC have mainly focused on using WET files while applying cleaning such as document level filtering \citep{GPT3,wenzek2019ccnet}, or n-sentence level deduplication with very aggressive heuristics \citep{T5}.

We do not believe that document level filtering is sufficient for WET files because many of the issues with WET files stem from intra-document boilerplate. We also find many of the heuristics used in \citet{T5}, such as the removal of all lines without terminal punctuation, the word "javascript", and 3-sentence deduplication to be too aggressive.

\subsubsection{Extraction}

In addition to jusText, we also considered Trafilatura, Newspaper, Goose3, and DragNet. While we were originally intending on creating an extraction benchmark, this proved infeasible given our available resources, and we chose jusText based on visual inspection of the output. In inspection, we noticed that jusText has the characteristic that it discards more data than many other extractors, which is not a major drawback given the large volume of CC data available. This was as expected, given jusText's intended application for text corpora creation. In contrast, trafilatura is, for instance, better at preserving the structure of the website faithfully, often correctly extracting elements such as tables, but it kept too much unnecessary boilerplate. Had we used trafilatura, we would have required an additional intra-page filtering step to remove boilerplate from the page.

\subsubsection{Languages}

While jusText does technically support several other languages, the quality on those languages is worse than on English as many constants in the algorithm are specifically tuned for English. Additionally, jusText is completely unable to handle languages such as Chinese and Japanese, which do not use spaces to delimit words.

Due to the difficulty of maintaining an acceptable level of extraction quality across all languages, we decided to restrict the scope of the CC dataset to only English and leave a high-quality, fully multilingual, WARC-based CC-based dataset to future work. To filter for only English, we use the pycld2 library and only attempt to extract text from documents where English is the most common language. 

We use pycld2 instead of fasttext because it is capable of classifying the language from the HTML directly, and since jusText requires knowledge of the language of the webpage before extraction. Additionally, pycld2 was significantly faster than jusText, and by only processing with jusText documents classified as English by pycld2, we reduced the required computation by approximately half.

Extracting text from websites for language modeling, especially for multilingual corpora, is highly nontrivial, and we leave the refinement of such extraction to future work. 

\subsubsection{Filtering}

\begin{figure*}%
    \centering
    \subfloat[\centering OpenWebText2]{{\includegraphics[width=6cm]{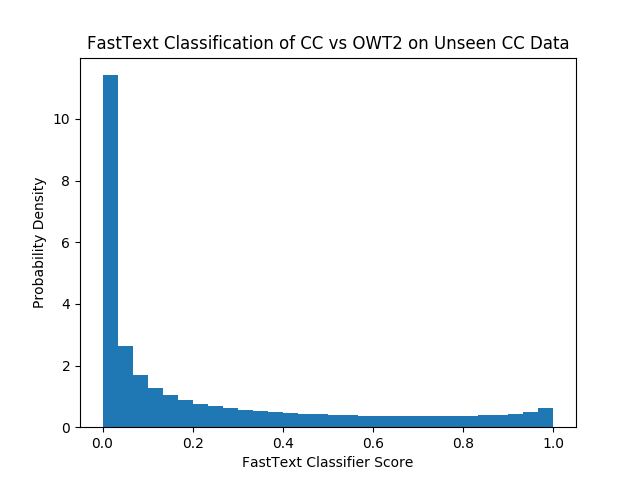} }}%
    \qquad
    \subfloat[\centering Full Pile]{{\includegraphics[width=6cm]{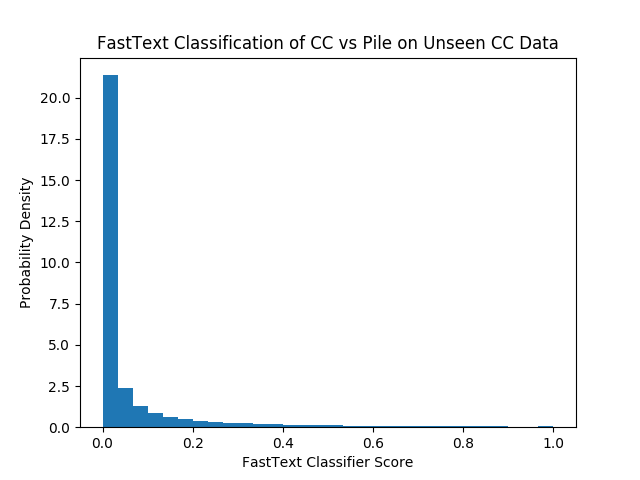} }}%
    \caption{Score distribution of documents from Common Crawl given different classifier training data.}%
    \label{fig:scores_ccvs}%
\end{figure*}

To filter CC for quality, we follow \citet{GPT3} in training a classifier to classify between a known high quality dataset and CC. We use fasttext with an n-gram size of 2. We ran experiments using both the entire Pile and just OpenWebText2 as the positive examples, with score distributions on unseen CC data as shown in Figure \ref{fig:scores_ccvs}. We decided to use only OpenWebText2 for positive examples for our final CC data because of the low sensitivity of using the full Pile. We use the same Pareto-distribution thresholding as \citet{GPT3}, with $\alpha = 3$. Our choice of $\alpha$ targets the filtering ratio necessary to filter our subset of CC to the size we needed. The impact of $\alpha$ on the filtering ratio is shown in Table \ref{table:cc_alpha_ratio}.

\begin{table}
\centering
\begin{tabular}{c c} 
\toprule 
$\alpha$ & Filtering Ratio \\ [0.5ex] 
\midrule 
1 & 0.5894 \\
2 & 0.3649 \\
3 & 0.2390 \\
4 & 0.1671 \\
5 & 0.1239 \\
6 & 0.0974 \\
7 & 0.0802 \\
8 & 0.0685 \\
9 & 0.0602 \\
\bottomrule 
\end{tabular}
\caption{Filtering Ratios (kept:total) of various settings} 
\label{table:cc_alpha_ratio} 
\end{table}

\subsection{Pubmed Central}

We use {\tt pandoc 1.19.2.4} \citep{pandoc} to convert the JATS format data provided by PMC to markdown. Afterwards, we remove any line beginning with {\tt :::}, which is used by pandoc to indicate html classes in markdown.

\subsection{Books3}

No additional details.

\subsection{OpenWebText2}

To produce the dataset, URLs and their associated metadata were first extracted from all Reddit submissions up to April 2020. URLs were deduplicated, with each unique URL featuring a list of associated submissions metadata, and an aggregate score. URLs with an aggregate score of less then 3 were removed. The links were then scraped and processed with Newspaper scraper. Deduplication was performed at the document level using in memory MinHashLSH through the DataSketch library.

Both filtered and raw versions were produced, with the raw version only deduplicated by URL. The filtered version contains 65.86 GB of uncompressed text across 17,103,059 documents. The raw version is much larger, at 193.89GB of uncompressed text across 69,547,149 documents.

\subsubsection{Extractor Choice}

We chose to use Newspaper instead of jusText for OpenWebText2 for consistency with OpenWebTextCorpus. Additionally, by using multiple different html extractors for different components of the Pile, we reduce the potential impact of systematic biases from any one extractor negatively impacting the dataset.

\subsection{ArXiv}

We downloaded the \TeX\ sources of all papers on arXiv up to the July 2020 dump (the last file included in our data is {\tt arXiv\_src\_2007\_068.tar}) via arXiv's S3 Bulk Source File Access\footnote{\url{https://arxiv.org/help/bulk_data_s3}}, and used {\tt pandoc 1.19.2.4} to convert these source files to Markdown, discarding any papers which had errors during the conversion process. This yielded a total of 1,264,405 papers.

We remove any line beginning with {\tt :::}, which is used by pandoc to indicate html classes in markdown.

\begin{figure}[ht]
  \includegraphics[width=\linewidth]{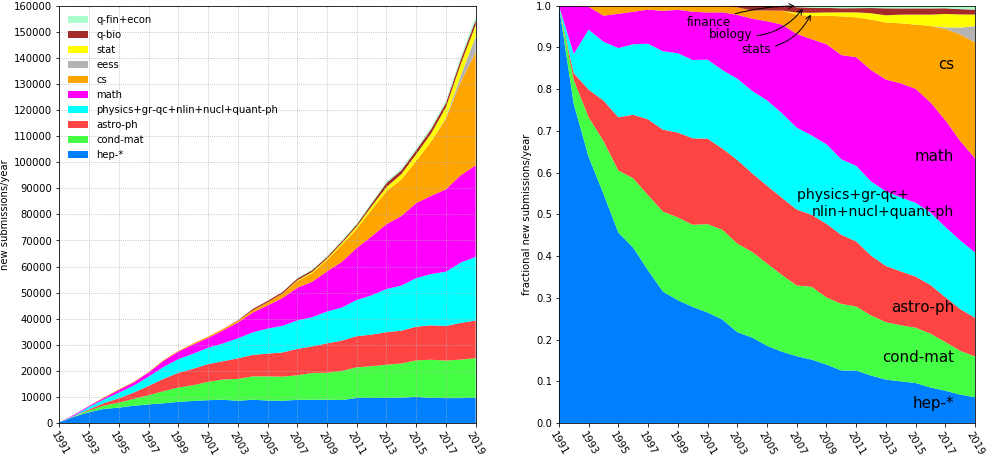}
  \caption{Left: number of new submissions/year to arXiv grouped by domain over time. Right: fractional submission rates for each of the domains.\protect\linebreak Figure from \url{https://arxiv.org/help/stats/2019_by_area/}}
  \label{fig:arXiv} 
\end{figure}

\subsection{GitHub}

We separate the data gathering process into two steps:

\begin{enumerate}  
\item Gathering a list of the desired repositories and their metadata
\item Extracting all text data useful for language modeling from each repository
\end{enumerate}


For the first step, mirroring the approach of the WebText dataset, we use GitHub ‘stars’ as a proxy for quality, and choose to gather only repositories with more than $100$ stars. For practical reasons, we also limit the list of repositories gathered to repositories with less than 1GB of files. Since Github’s API limits the number of search results to $1000$, in order to comprehensively gather all repositories we need to create many small queries that each return fewer than 1000 results in such a way that every repository of interest will be returned by at least one of our queries. To achieve this, we bound our initial search by size to return only repositories between a lower bound of $0$ and $5$ bytes. At the time of writing, this returns $965$ results. For the next step, we set our lower bound one above our previous upper bound, and decide on a new upper bound that should also return fewer than 1000 results by using the results from our last query to estimate our new upper bound as $(\mathrm{lower bound} + (1000 / (n / r))$, where $n$ is the number of previous results and $r$ is the range of bounds in the previous step.

This tends not to overshoot, because Github repositories follow a power distribution with respect to size, but if it does, we simply use the amount of repositories our new query returned in order to construct a new upper bound estimate.

Using the gathered list of repositories, we clone each one, extract any text-based files, and discard the rest. Because some repositories took an impractical amount of time to clone and/or extract, we set a hard time limit of 300 seconds for both the git cloning and text extraction steps. As such, some larger repositories may only be partially extracted. We also impose a file size limit of 100kB on extracted files, as we found that the majority of files over that size were typically very repetitive autogenerated source files or data files, and that setting this file size limit was an effective cleaning step to limit the data to code.

Because we wanted to limit the size of the overall Pile, we randomly sampled 95.0 GiB of the 630.64 GiB of Github data we collected in total and leave quality filtering to future work. 

However, we believe code generation will be an increasingly important component of language models as they continue to scale up and increase in their ability to generalize. As such, we hope to extend this dataset in future work.

\subsection{FreeLaw}

We download the court opinions data in bulk from CourtListener,\footnote{ \url{https://www.courtlistener.com/api/bulk-info/}} and extract the raw text using \texttt{BeautifulSoup}.

\subsection{Stack Exchange}

To construct the dataset, we download and parse every Stack Exchange database dump to plaintext files. We opt to extract the top three answers with at least three upvotes, discarding all other responses. We only include the plain text question and response and do not incorporate any metadata. Motivated by large-scale language models’ few-shot ability \citep{GPT3}, we provide context by prepending all questions and answers with \texttt{Q:\textbackslash n\textbackslash n} and \texttt{A:\textbackslash n\textbackslash n} respectively. 

The resulting dataset contains a total of 15,622,475 documents across a total of 365 Stack Exchanges and Meta-Stack Exchanges, the bulk of which is from StackOverflow.

\subsection{USPTO Backgrounds}

The United States Patent and Trademark Office (USPTO) has published bulk archives of the full text of all patents granted in the US from 1976 to September 2020. From these archives, we extract the Background sections, along with key grant-specific metadata, such as the inventor, assignee, and classification information.

The file format used for storing bulk text US patents has changed over time. Prior to 2002, all of the datasets are in a specialized format called APS (Automated Patent System). Since 2002, the data is XML encoded. Partially as a function of this change, the location of the "Background" section has also shifted. Our converter accounts for these structural shifts and extracts the raw text from each patent's Background.

\subsection{PubMed Abstracts}

About one-third of the articles in the dataset were missing or contained a malformed title or abstract and were excluded. Additionally, PubMed Central (see Section \ref{subsection:pmc}) contains full-text resources to many recent publications; any publications which already appear in PMC are excluded from this set. To process the data, we concatenated the title and abstract and removed any copyright information. The remaining dataset contains 15,518,009 titles and abstracts.

\subsection{Project Gutenberg}

No additional details.

\subsection{OpenSubtitles}

To create the text dataset, we simply extract the subtitle text from each XML file in the English language dataset provided by \citet{OpenSubtitles}, discarding any provided metadata.

\subsection{Wikipedia (English)}

We use the \texttt{wikipedia/20200301.en} dataset from TensorFlow Datasets.\footnote{\url{https://www.tensorflow.org/datasets/catalog/wikipedia\#wikipedia20200301en}} We prepend the title to the body of each article, separated by two newlines.

\subsection{DeepMind Mathematics}

We include instances from the Easy, Medium, and Hard components of DeepMind Mathematics, breaking each curriculum item (such as {\tt algebra\_\_polynomial\_roots}) into 8 KiB chunks.

\subsection{Ubuntu IRC}

We processed all logs from July 5, 2004 through September 1, 2020.

To process the data, all system messages, such as joins, disconnects, nick changes, etc. were discarded, but actions (i.e using {\tt /me}) were kept. Timestamps were removed, and all logs for the same channel in a given week were concatenated into a single document, with each the logs for each day prepended with the date if that day's log is non-empty.

\subsection{BookCorpus2}

The original BookCorpus consists of 11,038 books. However, due to issues with availability of the original BookCorpus, as well as the possibility of collecting a larger version, we decided to collect our own version of BookCorpus using a similar methodology as \citet{BookCorpusCode}. Our version of BookCorpus contains 17,868 books instead.

We create and use a modified version of the epub-to-text converter in \citet{BookCorpusCode} that:

\begin{itemize}
  \item Correctly preserves the document structure across chapters, matching the table of contents very closely;
  \item Correctly renders tables of data, whereas by default \texttt{html2txt} produces poor-quality results for tables,
  \item Correctly preserves code structure, so that source code is visually coherent,
  \item Converts numbered lists from ``1\textbackslash.'' to ``1.''
  \item Runs the full text through {\tt ftfy.fix\_text()} \citep{speer-2019-ftfy}, replacing Unicode apostrophes with ascii apostrophes and expanding Unicode ellipses to ``...'' (three separate ascii characters).
\end{itemize}

\subsection{EuroParl}

We download the data in bulk from \footnote{\url{ http://www.statmt.org/europarl/}}. We remove all basic tag information and only retain the name of each document as a title. For example, {\tt <SPEAKER ID=77 LANGUAGE="NL" NAME="Pronk">} becomes {\tt Pronk}, and then extract the body of each document, discarding those that are shorter than 200 characters.

\subsection{HackerNews}
We first use the Hackernews BigQuery dataset to obtain a list of all story ids in our date range. For the Pile we use the first Hacker News post (1) to post number 24531712. This corresponds to a date range of approximately 10/09/2006 to 09/20/2020. We use the BigQuery dataset to gather story ids for efficiency purposes. However, the BigQuery dataset was lacking some information for stories, so we used the official Hacker News API for story and comment text retrieval. 

Hacker News displays and stores comments in a tree-like manner, with children comments replying to parent comments. However, most language models require input data to be in a sequential form. Considering each path through the comment tree as a sequence could be detrimental, since there will be a large amount of near-duplicate comment sequences. In addition, only taking one path through the comment tree for each story leaves out a large portion of the comment data. Therefore, we parsed comments in a hybrid form. For every top-level comment (comments that have no parent comment), we create a sequence of comments by traversing down the comment tree from the top-level comment. We choose the next comment by taking the child comment with the highest number of children comments (a cheap attempt at taking a long path through the comment tree, note that it does not take the longest possible path).

We consider all stories that have at least one comment and are not flagged by the moderators for potential conduct violations. Since comments are stored in HTML, we use the {\tt html2text} package to extract the text from the post.

We order each document by listing the title, url, sub-title, and author at the top. Top-level comments are delimited by "\texttt{\textbackslash n------\textbackslash n}" and sub-comment chains are delimited by "\texttt{\textbackslash n\textasciitilde{}\textasciitilde{}\textasciitilde{}\textbackslash n}". We include author and extracted text for each comment.

\subsection{YouTube Subtitles}

We construct the dataset in three stages:

\begin{enumerate}  
\item We build a large list of search terms by prompting a GPT-3 model with a manually selected list of queries, manually filtering the responses, and repeating this process iteratively until a suitable size is reached. The list of terms is centred around, but not limited to, educational topics.
\item We use requests-html to gather a list of 1000 Youtube video IDs for each search term, and deduplicate the resulting video ids across search terms.
\item We use \texttt{YoutubeTranscriptApi}\footnote{ \url{https://github.com/jdepoix/youtube-transcript-api}} to gather all human generated closed captions for every available language for each video. To align each language in parallel, we split the captions for each language into parallel minute-long sections by timestamp, and arrange each language in a random order within these sections, appending the language as a header to each minute-long section to provide context. If only a single language is available, the output is just the subtitles, with no header appended.
\end{enumerate}

In total, subtitles for 173,651 videos were gathered.

\subsection{PhilPapers}

The PhilPapers (PP) are indexed using OAI-MPH, the Open Archives Initiative Protocol for Metadata Harvesting. As such, the first step to collect the data is to get the XML for all links. This was done using \texttt{pyoaiharvester}.\footnote{\url{https://github.com/vphill/pyoaiharvester/}}

From that, each publication is downloaded. Some entries do not exist, or have been removed by the authors. Papers with text are extracted using \texttt{pdfbox}, and papers with non-machine readable text are ignored. Non-English language publications are kept, and the metadata reflects the language reported by the OAI-MPH XML. The text is filtered with \texttt{pdf\_filter.py} from PDFextract, and we discard any papers with less than 1000 characters.\footnote{\url{https://github.com/sdtblck/PDFextract}}

\subsection{NIH Grant abstracts: ExPORTER}

The NIH provides a bulk-data repository for awarded applications through the ExPORTER service covering the fiscal years 1985--present. These data come from the NIH, but also other other Health and Human Services agencies (ACF, AHRQ, CDC, HRSA, FDA), and the VA. Additionally, the NIH provides a legacy data format named CRISP for awarded applications during the fiscal years 1970--2009.

We merged both the ExPORTER and CRISP data to form a consolidated dataset of awarded applications. Entries were deduplicated based off their application ID, and excluded if their abstract text was missing or too short. Small grants, especially administrative ones, consisted solely of short boilerplate. For this reason, we further deduplicated on abstract text. All grants types were considered, including new applications (Application Type Code 1) and renewals (Application Type Code 2) as the text differed enough to provide novel input. The text was then minimally parsed to remove administrative boilerplate, (ex. most old awards contain some variation of ``description: (provided by applicant)"). In total, there were 939,668 grant application abstracts added.

\subsection{Enron Emails}

To extract the data, we used the {\tt mailparser} package\footnote{\url{https://github.com/SpamScope/mail-parser}} to extract the body of each email as a document.

\section{General Data Processing}
This section discusses any processes applied across multiple datasets.


To combine the constituent datasets, we iterate until the size of the output dataset is the desired size, drawing documents from datasets at random, weighted by the number of documents in each dataset times the number of epochs desired on that dataset. Because the number of documents involved is high, by the law of large numbers, the number of copies of each dataset present in the Pile is approximately equal to its epoch count.

Shuffling a dataset posed a major problem due to our limited memory and computational budget. We follow \citet{hardin_2018}, a method descended from \citet{10.2307/25049166}, and interleave our output to produce 30 output piles. 

We hold out approximately 10GiB of data from the Pile, of which 2GiB are used to create the validation and test splits, and the remainder is held in reserve. From the training set, we remove any elements that are also present verbatim in any of the held out data, to prevent leakage.

\subsection{Weights}\label{apdx:methods}

Similar to \citet{GPT3}, we increase the weight of certain components such that the number of epochs elapsed on data we consider high quality is greater than one. Our choice of weights was primarily informed by the source of the data and the size of the dataset; we attempted to upweight academic texts the most, which we felt provided the highest quality data, as well as smaller sets, such that they would have a more pronounced impact on the data. We strictly disallowed any data more than 3 epochs and avoided having any data with more than 2 epochs.

\subsection{Deduplication}\label{apdx:deduplication}


Due to memory constraints we did not perform Pile wide de-duplication. Instead, de-duplication was performed at the document level within OpenWebText2 and Pile-CC as those sets were the most likely to contain duplicate documents. 

The same technique was used for both OpenWebText2 and Common Crawl---MinHashLSH with the Python Datasketch library.\footnote{\url{https://github.com/ekzhu/datasketch}} We used 10 hash functions for each Minhash and an approximate Jaccard similarity of 0.5. This produced a duplicate rate of 28\% in OpenWebText2 and 26\% for Common Crawl.

The main challenge here was computational, leading us on a journey through the various LSH persistence options. A simple quadratic Minhash comparison of all documents would have taken several hundred thousand years, motivating the use of LSH. Initially, we did not have sufficient RAM for in-memory LSH and chose to use the Cassandra backend when de-duplicating OpenWebText2. This was reasonably fast, but the same method resulted in a corrupted database about $\frac{3}{4}$ of the way through processing Common Crawl. After the Cassandra corruption, we briefly tested the experimental Mongo implementation; however this was quite slow due to the nature of Mongo itself. In the end, we ran in-memory LSH on a machine with enough RAM for Common Crawl, taking several days.

\subsection{Downstream Validation Leakage}

To avoid leakage of data from downstream evaluations, recent work \citep{GPT2,GPT3,Megatron} has removed any data in the training set that may overlap with the evaluation metrics. We decided not to perform any such removal, because it is impossible to anticipate all potential downstream evaluation metrics, and so any particular selection of metrics would inevitably either become obsolete as the choice of benchmarks in the field changes, or potentially hinder the development of new benchmarks for models trained on Pile.

For models trained on Pile and evaluated on metrics other than Pile's own validation and test sets, we encourage authors to remove overlaps between Pile and the validation data of these additional downstream evaluations. We do not anticipate that such leakage removal will hurt model performance, as the validation sets of most benchmarks are very small in relation to the size of the Pile, and so choosing to evaluate on more metrics will not be a disadvantage for any model.

\begin{table}[t]

    \centering
    \begin{tabular}{l r}
    \toprule
        Component & Tokens per byte \\ & ($L_T/L_B$)
        \\
        \midrule
Pile-CC & 0.2291 \\
PubMed Central & 0.3103 \\
Books3 & 0.2477 \\
OpenWebText2 & 0.2434 \\
Arxiv & 0.3532 \\
Github & 0.4412 \\
FreeLaw & 0.2622 \\
StackExchange & 0.3436 \\
USPTO Backgrounds & 0.2116 \\
PubMed Abstracts & 0.2183 \\
Gutenberg (PG-19) & 0.2677 \\
OpenSubtitles & 0.2765 \\
Wikipedia (en) & 0.2373 \\
DM Mathematics & 0.8137 \\
Ubuntu IRC & 0.3651 \\
BookCorpus2 & 0.2430 \\
EuroParl & 0.3879 \\
HackerNews & 0.2627 \\
YoutubeSubtitles & 0.4349 \\
PhilPapers & 0.2688 \\
NIH ExPorter & 0.1987 \\
Enron Emails & 0.3103 \\
        \bottomrule
    \end{tabular}

\caption{Tokens per byte for Pile components}
\label{tbl:bpb_conversion}
\end{table}

\section{Investigating data}
\subsection{13-Gram Analysis}


As part of our exploratory analysis, we calculated the counts of all 13-grams across Common Crawl. We chose $n=13$ due to its use in prior work \citep{GPT3}. There were a total of 40,216,231,078 different 13-grams in this dataset. The 1000 most common range from 11 million occurrences down to 20k.

The most frequently occurring 13-grams were character repetitions used for styling such as ``\texttt{-{}- -{}-}'', ``\texttt{* * * *}'', ``\texttt{! ! ! !}'', at 11 million, 5.8 million and 1.1 million respectively. Other characters used in this manner include the following: ``\texttt{\# . > ?}''. In the 264k count range, we see repetitions of badly formatted HTML escape characters ``\texttt{; \&nbsp}'', ``\texttt{; amp}''. Boilerplate from standard forum software appears around the 180k occurrences range, such as the following: ``\texttt{select the forum that you want to visit from the selection below}''.

Overall, a large amount of common HTML and CSS is included in the top 1000, along with boilerplate text from Amazon Affiliate Advertising, TripAdvisor, SimplyHired, Associated Press, PostMedia, The FCC etc. PHP error messages and password login prompts also made an appearance. It may be of interest to fans of Portal that repetitions of ``\texttt{the cake is a lie .}'' achieved a high count. 

\subsection{Benchmark Perplexity Computation}
\label{apdx:perplexity}

To compute the perplexity for a given dataset, we tokenize each document separately, divide the document into segments of up to the maximum sequence length of the model (1024 tokens for GPT-2, 2048 for GPT-3), and predict the logits of the each segment.
The inputs to the model are the immediate prior tokens the e.g. for scoring tokens 1 to 1024, we provide tokens 0 to 1023 at the input context.
The respective language model implementations handle the causal attention masking.
This ensures that every token in the dataset is scored exactly once.
This also means that some tokens will have more input context than others.
We then aggregate over the whole dataset and compute the final perplexity score.
The perplexity for the whole Pile is computed by aggregating over the constituent datasets (i.e. weighted by dataset size, not a simple average of dataset perplexities).
Both GPT-2 and GPT-3 share the same tokenizer and vocabulary, making the perplexity scores directly comparable. 
We use the Hugging Face \citep{wolf2020transformers} implementation of GPT-2, and the OpenAI API for GPT-3. The \texttt{davinci} model in the OpenAI API is presumed to correspond to a 175B parameter version of GPT-3.

In Table~\ref{table:pile_perplexity} we show the test set perplexities (i.e. not normalized by UTF-8 length, as in Table~\ref{table:pile_perplexity_utf8}). 
Because of the costs associated with using the OpenAI API, we compute test perplexities on only one-tenth of the test set in Tables~\ref{table:pile_perplexity} and Table~\ref{table:pile_perplexity_utf8}. Specifically, we randomly sample one-tenth of the documents of each dataset except for three: Ubuntu IRC, BookCorpus2, and PhilPapers. In Table~\ref{table:pile_perplexity_fullgpt2}, we show test perplexity computed on the full test set on all GPT-2 models.

\begin{table*}[p]
\resizebox{\textwidth}{!}{\small%
    \centering
    \begin{tabular}{l rrrr rrrr}
    \toprule
        \textbf{Component} 
        & \multicolumn{4}{c}{GPT-2}
        & \multicolumn{4}{c}{GPT-3}
        \\ \cmidrule(lr){2-5} \cmidrule(lr){6-9}
        & \multicolumn{1}{c}{small}
        & \multicolumn{1}{c}{medium}
        & \multicolumn{1}{c}{large}
        & \multicolumn{1}{c}{xl}
        & \multicolumn{1}{c}{ada}
        & \multicolumn{1}{c}{babbage}
        & \multicolumn{1}{c}{curie}
        & \multicolumn{1}{c}{davinci}
        \\
        \midrule
    Pile-CC & 26.8894 & 20.5671 & 18.1656 & 16.9572 & 16.2430 & 13.0270 & 10.7532 & 8.4929 \\
    PubMed Central & 11.0626 & 8.9052 & 8.0454 & 7.5404 & 6.8800 & 5.7006 & 4.9390 & 4.3143 \\
    Books3 & 28.3889 & 22.0958 & 19.3424 & 17.7833 & 15.4209 & 12.4220 & 10.1526 & 7.1927 \\
    OpenWebText2 & 23.6764 & 17.6175 & 15.1314 & 13.6267 & 12.0063 & 9.5439 & 7.7706 & 5.9163 \\
    ArXiv & 14.2804 & 11.1896 & 10.0904 & 9.3330 & 7.5551 & 6.1541 & 5.2537 & 4.5341 \\
    Github & 16.6814 & 7.9322 & 16.6742 & 13.3337 & 3.9614 & 3.1660 & 2.7398 & 2.4240 \\
    FreeLaw & 16.1000 & 11.7518 & 10.8427 & 10.0965 & 8.7976 & 7.0366 & 5.8256 & 4.8926 \\
    Stack Exchange & 13.7202 & 9.3405 & 8.8467 & 8.3238 & 7.6652 & 5.9486 & 5.0267 & 4.3796 \\
    USPTO Backgrounds & 15.1141 & 11.9232 & 10.5878 & 9.8095 & 9.2775 & 7.7000 & 6.5849 & 5.6411 \\
    PubMed Abstracts & 20.5642 & 15.2379 & 13.1190 & 11.9355 & 13.2112 & 10.4188 & 8.5861 & 7.1604 \\
    Gutenberg (PG-19) & 26.4947 & 17.8975 & 16.4722 & 16.5112 & 12.5709 & 9.6349 & 7.7940 & 6.3112 \\
    OpenSubtitles & 22.7418 & 18.5724 & 17.0868 & 16.2709 & 16.2174 & 13.8561 & 11.8836 & 9.8578 \\
    Wikipedia (en) & 27.0237 & 19.7570 & 17.4856 & 16.7849 & 12.9112 & 9.9453 & 7.8363 & 5.6915 \\
    DM Mathematics & 9.8990 & 8.7389 & 8.2928 & 7.9772 & 7.2458 & 6.5231 & 6.0171 & 5.6020 \\
    Ubuntu IRC & 33.3028 & 26.1203 & 22.6128 & 20.9461 & 12.1138 & 9.6995 & 8.0628 & 6.5679 \\
    BookCorpus2 & 25.0743 & 19.9725 & 17.6343 & 16.2905 & 16.1530 & 13.1796 & 11.0885 & 9.2205 \\
    EuroParl & 62.8981 & 36.9757 & 28.6198 & 23.4294 & 6.4996 & 5.3282 & 4.4982 & 3.8327 \\
    HackerNews & 45.0915 & 29.2599 & 32.0796 & 33.9774 & 22.1295 & 17.6314 & 14.6582 & 12.1283 \\
    YoutubeSubtitles & 25.7794 & 18.8173 & 15.9002 & 14.3104 & 8.4740 & 6.6394 & 5.4510 & 4.5235 \\
    PhilPapers & 30.1129 & 23.0288 & 20.3755 & 18.5649 & 14.4730 & 11.6785 & 9.6797 & 7.9915 \\
    NIH ExPorter & 23.9004 & 18.2298 & 15.9850 & 14.6371 & 16.1417 & 12.8744 & 10.6573 & 8.8110 \\
    Enron Emails & 34.7954 & 23.4353 & 25.7138 & 23.9791 & 16.8190 & 13.6043 & 11.6473 & 9.7655 \\
    \midrule
    The Pile & 18.0878 & 13.2253 & 12.9177 & 11.8633 & 9.7355 & 7.8456 & 6.5904 & 5.4508 \\
        \bottomrule
    \end{tabular}
}
\caption{Test perplexity of the Pile using GPT-2 and GPT-3. Evaluation is performed on one-tenth of the test data of the Pile, on a per-document basis.}
\label{table:pile_perplexity}
\end{table*}

\begin{table*}[ht]
    \centering
    \begin{tabular}{l rrrr}
    \toprule
        \textbf{Component} 
        & \multicolumn{4}{c}{GPT-2}
        \\ \cmidrule(lr){2-5}
        & \multicolumn{1}{c}{small}
        & \multicolumn{1}{c}{medium}
        & \multicolumn{1}{c}{large}
        & \multicolumn{1}{c}{xl}
        \\
        \midrule
    Pile-CC & 26.5 & 20.3 & 17.9 & 16.7  \\
    PubMed Central & 10.3 & 8.3 & 7.6 & 7.1  \\
    Books3 & 27.9 & 21.3 & 18.5 & 17.0  \\
    OpenWebText2 & 23.5 & 17.5 & 15.1 & 13.6  \\
    ArXiv & 13.7 & 10.7 & 9.7 & 9.0  \\
    Github & 16.5 & 8.1 & 16.7 & 13.5  \\
    FreeLaw & 15.9 & 11.6 & 10.8 & 10.1  \\
    Stack Exchange & 13.7 & 9.4 & 9.0 & 8.4  \\
    USPTO Backgrounds & 16.6 & 13.0 & 11.5 & 10.6  \\
    PubMed Abstracts & 20.9 & 15.4 & 13.3 & 12.1  \\
    Gutenberg (PG-19) & 37.8 & 24.9 & 22.8 & 24.3  \\
    OpenSubtitles & 22.1 & 18.1 & 16.6 & 15.8  \\
    Wikipedia (en) & 27.0 & 19.8 & 17.5 & 16.8  \\
    DM Mathematics & 9.9 & 8.7 & 8.3 & 7.9  \\
    Ubuntu IRC & 33.3 & 26.1 & 22.6 & 20.9  \\
    BookCorpus2 & 25.1 & 20.0 & 17.6 & 16.3  \\
    EuroParl & 63.9 & 41.9 & 33.5 & 27.7  \\
    HackerNews & 43.7 & 28.3 & 30.9 & 32.4  \\
    YoutubeSubtitles & 25.3 & 18.8 & 16.2 & 14.8  \\
    PhilPapers & 30.1 & 23.0 & 20.4 & 18.6  \\
    NIH ExPorter & 23.2 & 17.7 & 15.5 & 14.2  \\
    Enron Emails & 22.0 & 15.4 & 18.6 & 18.1  \\
    \midrule
    the Pile & 18.4 & 13.3 & 13.1 & 12.0 \\
        \bottomrule
    \end{tabular}
\caption{Full Test Perplexity of the Pile using GPT-2.}
\label{table:pile_perplexity_fullgpt2}
\end{table*}

\begin{figure}[ht]
  \includegraphics[width=\linewidth]{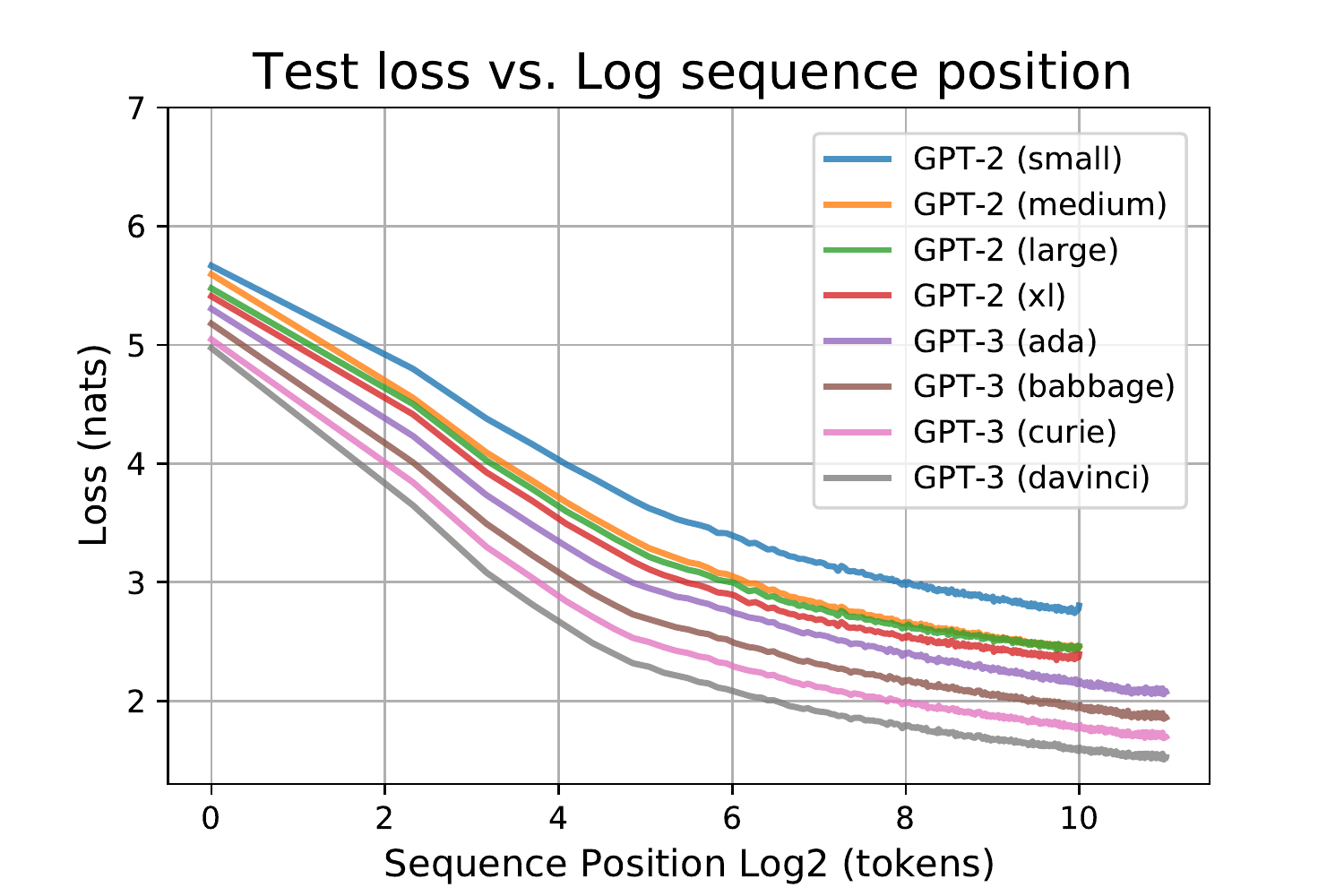}
  \caption{Test loss (log perplexity) over the Pile, bucketed by position in the input sequence based on the model's maximum sequence length. To smooth out the lines, we bucket 4 positions per plotted datapoint. (e.g. positions 0--3, positions 2044--2047). Later tokens are predicted with more context and thus see lower perplexities.} 
   \label{fig:position_perplexity}
\end{figure}

\subsection{Pejorative Content}\label{apdx:own_pejorative}

\begin{figure}[ht]
  \includegraphics[width=\linewidth]{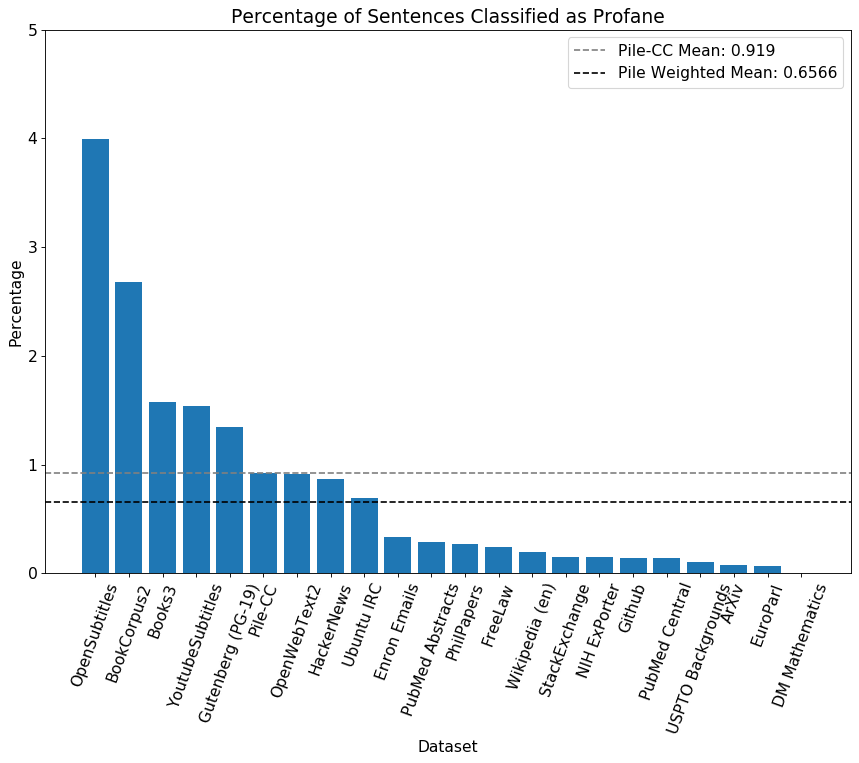}
  \caption{Percentage of sentences classified as profane in the Pile. The percentage of the CC component and the weighted mean of the Pile as a whole are shown as horizontal lines} 
   \label{fig:profane_sent}
\end{figure}

Initially we decided on separating pejorative content into 4 groups: sex-related terminology, slurs, neither of these categories, and both of these categories. We adapted a public "naughty words" list and broke them into these categories with the intern of looking at the proportion of each category in each dataset. However, this provided many issues. 

First, any blacklist of words would be hard-pressed to catch all the instances of pejorative content, since purposeful misspellings of words could evade the censor and still have the intended effect. Furthermore, words and their intents are always evolving, therefore any list created would likely be always outdated. 
Another issue pertains to sorting the words into the categories. Words are highly dependent on their context, so a word would change categories with different contexts.

\section{Data Samples}

The following consists of two random, non-cherrypicked 512-byte samples from each constituent dataset of the Pile, sampled from the validation split.

\subsection{Pile-CC}
\centerline{\rule{0.9\linewidth}{1pt}}
\begin{tiny}\begin{quote}
pot trending topics and the coverage around them. First up, there’s a bit of a visual redesign. Previously, clicking on a trending topic would highlight a story from one publication, and you’d have to scroll down past a live video section to view related stories. Facebook is replacing that system with a simple carousel, which does a better job of showing you different coverage options. To be clear, the change doesn’t affect how stories are sourced, according to Facebook. It’s still the same algorithm pickin
\end{quote}\end{tiny}

\centerline{\rule{0.9\linewidth}{1pt}}
\begin{tiny}\begin{quote}
e public safety. He said the bridge saves commuters two or three minutes when trains pass – and those minutes could be vital.\\
\\
“Two to three minutes may not mean much if you’re just driving home from work, but if you’re the one waiting for an ambulance to get to your home, if you’re the one waiting for a fire truck to get to your home, if you’re the one waiting for a police car to get to your home, those two to three minutes could mean the difference between life or death,” Sharp said. “That’s what this pro
\end{quote}\end{tiny}
\centerline{\rule{0.9\linewidth}{1pt}}

\subsection{PubMed Central}
\centerline{\rule{0.9\linewidth}{1pt}}
\begin{tiny}\begin{quote}
d a suitable substitute for the advice of a qualified health care professional. Do not disregard or avoid professional medical advice due to content published within Cureus.\\
\\
Introduction\\
============\\
\\
Total knee arthroplasty (TKA) is a promising treatment for end-stage osteoarthritis (OA) of the knee for alleviating pain and restoring the function of the knee. Some of the cases with bilateral TKA are symptomatic, necessitating revision arthroplasty in both the knees. A bilateral revision TKA can be done ei
\end{quote}\end{tiny}

\centerline{\rule{0.9\linewidth}{1pt}}
\begin{tiny}\begin{quote}
ent\textbackslash{}'s ability to make judgements and decisions about their work experiences and learning that will position them as future critical thinkers, life longer enquirers and learners.\\
\\
Conclusion \{\#jmrs290-sec-0014\}\\
==========\\
\\
Identification of the core capabilities that our stakeholder community rate highly has proved informative in assisting us to describe a "work ready *plus"* medical imaging graduate for the New Zealand context. The results have provided data to the curriculum development team allowing them
\end{quote}\end{tiny}
\centerline{\rule{0.9\linewidth}{1pt}}

\subsection{Books3}
\centerline{\rule{0.9\linewidth}{1pt}}
\begin{tiny}\begin{quote}
cept of \_forçage\_ , 'a forcing of language enacted by the advent of an "other" language that is at once immanent and created', 44as Badiou puts it: this opens up vistas of a truly syntactic analysis of the poem, in which, again, Badiou would be close to his philosophical other, Deleuze, who, as we just saw, defines style through a-grammaticality and who tries to define what he calls an 'intensive line of syntax'.45\\
\\
Nevertheless, the insistence on syntax as guarantee involves a \_seventh paradox\_ , the parad
\end{quote}\end{tiny}

\centerline{\rule{0.9\linewidth}{1pt}}
\begin{tiny}\begin{quote}
rnment, before the Second World War there were 5,300 communities and two million burakumin. The BLL thinks there must be at least three million burakumin living in Japan today.\\
\\
We visited a hall in Osaka where a taiko drum group, made up exclusively of young burakumin, were about to start their weekly rehearsal. The small gymnasium was filled with taiko drums of all sizes. The smallest was about the size of a snare drum, the largest about the size of a compact car. The Japanese drum group Kodo have made th
\end{quote}\end{tiny}
\centerline{\rule{0.9\linewidth}{1pt}}

\subsection{OpenWebText2}
\centerline{\rule{0.9\linewidth}{1pt}}
\begin{tiny}\begin{quote}
prime minister to repatriate all the police sent to Catalonia before the referendum.\\
\\
What were the results?\\
\\
With nearly all votes counted, the pro-independence parties Together for Catalonia (JxCat), Republican Left of Catalonia (ERC) and Popular Unity (CUP) were on course to win a total of 70 seats in total, giving them a majority in the new parliament.\\
\\
Citizens (Cs) had 25.3\% of the vote, winning 37 seats in the 135-seat chamber.\\
\\
Its leader told the BBC her party had been "victorious". Ms Inés Arrima
\end{quote}\end{tiny}

\centerline{\rule{0.9\linewidth}{1pt}}

\begin{tiny}\begin{quote}
so accommodate Stablecoins.\\
\\
While some analysts opined that Stablecoins are created to bring growth into the crypto space, they are becoming a solid way to reduce crypto volatility due to the fact that their value are pegged to fiat currency.\\
\\
For Low Cost and Almost Instant Across Border Remittance\\
\\
When Stellar-based Wirex stablecoins finally launches, they are going to be used to perfect low-cost and all most immediately cross-border remittance, just like the IBM Stablecoin which has received support fr
\end{quote}\end{tiny}

\centerline{\rule{0.9\linewidth}{1pt}}

\subsection{ArXiv}

\centerline{\rule{0.9\linewidth}{1pt}}

\begin{tiny}\begin{quote}
\textbackslash{}mbox\{ if \} 2\textbackslash{}leq x\textbackslash{}leq 3\\
\textbackslash{}end\{array\}\textbackslash{}right.\$\$ is lower semicontinuous and the nonemptiness of \$\{\textbackslash{}operatorname\{fix\}\}\textbackslash{}Phi\$ is guaranteed by Corollary \textbackslash{}[cor:fixed point\textbackslash{}]. Notice that \$\{\textbackslash{}operatorname\{fix\}\}\textbackslash{}Phi=[1,2]\$. Nevertheless the Kakutani fixed point Theorem does not apply since \$\{\textbackslash{}operatorname\{gph\}\}\textbackslash{}Phi\$ is not closed.\\
\\
On the converse, the set-valued map \$\textbackslash{}Phi:[0,3]\textbackslash{}rightrightarrows [0,3]\$ \$\$\textbackslash{}Phi(x):=\textbackslash{}left\textbackslash{}\{\textbackslash{}begin\{array\}\{ll\}\\
\textbackslash{}\{1\textbackslash{}\} \& \textbackslash{}mbox\{ if \} 0\textbackslash{}leq x\textless{}1\textbackslash{}\textbackslash{}\\
\{\}[1,2] \& \textbackslash{}mbox\{ if \} 1\textbackslash{}leq x\textbackslash{}leq 2\textbackslash{}\textbackslash{}\\
\textbackslash{}\{2\textbackslash{}\} \&
\end{quote}\end{tiny}

\centerline{\rule{0.9\linewidth}{1pt}}

\begin{tiny}\begin{quote}
.eps)\{width="6.5cm"\}\\
\\
![Gamma-ray spectrum at Mt. Norikura (2.77 km a.s.l). The vertical axis is Flux\$\textbackslash{}times E\^{}2\$. Our data is at \$\textless{}\$ 100 GeV. Data above 300 GeV is from emulsion chamber experiments. For the latter, see Sec.\textbackslash{}[discuss\textbackslash{}] []\{data-label="norispec"\}](norikura.eps)\{width="7.5cm"\}\\
\\
![The altitude variation of the flux integrated over 6 GeV. The dpmjet3.03 and fritiof7.02 give almost the same feature consistent with the observation while the deviation of fritiof1.6 from the data is obvious. \textbackslash{}[trans
\end{quote}\end{tiny}

\centerline{\rule{0.9\linewidth}{1pt}}

\subsection{Github}

\centerline{\rule{0.9\linewidth}{1pt}}

\begin{tiny}\begin{quote}
"enabled", out.enabled);\\
\}\\
\\
std::string SMTPServerInfoJSONStringSerializer::serialize(const SMTPServerInfo \&in,\\
                                                          const SecurityContext \&sc)\\
\{\\
    return SMTPServerInfoJSONSerializer::serialize(in, sc).dump(4);\\
\}\\
\\
void SMTPServerInfoJSONStringSerializer::unserialize(SMTPServerInfo \&out,\\
                                                     const std::string \&in,\\
                                                     const SecurityContext \&sc)\\
\{\\
    retur
\end{quote}\end{tiny}

\centerline{\rule{0.9\linewidth}{1pt}}

\begin{tiny}\begin{quote}
lose"\textgreater{}\\
    \textless{}at-form state="vm.form" autocomplete="off" id="external\_test\_form"\textgreater{}\\
        \textless{}at-input-group col="12" tab="20" state="vm.form.inputs" form-id="external\_test"\textgreater{}\textless{}/at-input-group\textgreater{}\\
        \textless{}at-action-group col="12" pos="right"\textgreater{}\\
            \textless{}at-action-button\\
                variant="tertiary"\\
                ng-click="vm.onClose()"\\
            \textgreater{}\\
                \{\{::vm.strings.get('CLOSE')\}\}\\
            \textless{}/at-action-button\textgreater{}\\
            \textless{}at-action-button\\
                variant="primary"\\
                n
\end{quote}\end{tiny}

\centerline{\rule{0.9\linewidth}{1pt}}

\subsection{FreeLaw}

\centerline{\rule{0.9\linewidth}{1pt}}

\begin{tiny}\begin{quote}
ssible, and further, that the weight of the evidence, the credibility of the witnesses and the persuasive effect of the testimony is for the sole determination of the trier of fact.\\
This Court thus uses the same interpretation of V.R.C.P. 52(a) as it did *487 under the previous statutory requirement found in 12 V.S.A. § 2385.\\
In essense, the defendants urge that this Court should reconsider the case of Green Mountain Marble Co. v. Highway Board, supra, and follow the Federal practice of looking to the evide
\end{quote}\end{tiny}

\centerline{\rule{0.9\linewidth}{1pt}}

\begin{tiny}\begin{quote}
ng to the fact that\\
the relevant Arkansas statutes and rules provide for criminal sanctions against school\\
officials who fail to enforce the immunization requirements, the Morningstar and\\
Lake Hamilton School Districts characterized themselves as disinterested bystanders\\
caught in the crossfire between the Schoolchildren and the Officials. See Ark. Code\\
Ann. § 6-18-702(c)(2)(B) (2000) (“Any school official, parent, or guardian violating\\
the regulations shall be subject to the penalties imposed herein.”); Id
\end{quote}\end{tiny}

\centerline{\rule{0.9\linewidth}{1pt}}

\subsection{Stack Exchange}

\centerline{\rule{0.9\linewidth}{1pt}}

\begin{tiny}\begin{quote}
ooks like a fancy wheel, When resetting rotation from 360deg to 0 deg, It animating the wheel in anti-clockwise direction, How to Avoid this???\\
HTML\\
\textless{}ul class="cm"\textgreater{}\\
  \textless{}li\textgreater{}\textless{}span\textgreater{}01\textless{}/span\textgreater{}\textless{}/li\textgreater{}\\
  \textless{}li\textgreater{}\textless{}span\textgreater{}02\textless{}/span\textgreater{}\textless{}/li\textgreater{}\\
  \textless{}li\textgreater{}\textless{}span\textgreater{}03\textless{}/span\textgreater{}\textless{}/li\textgreater{}\\
  \textless{}li\textgreater{}\textless{}span\textgreater{}04\textless{}/span\textgreater{}\textless{}/li\textgreater{}\\
  \textless{}li\textgreater{}\textless{}span\textgreater{}05\textless{}/span\textgreater{}\textless{}/li\textgreater{}\\
  \textless{}li\textgreater{}\textless{}span\textgreater{}06\textless{}/span\textgreater{}\textless{}/li\textgreater{}\\
  \textless{}li\textgreater{}\textless{}span\textgreater{}07\textless{}/span\textgreater{}\textless{}/li\textgreater{}\\
  \textless{}li\textgreater{}\textless{}span\textgreater{}08\textless{}/span\textgreater{}\textless{}/li\textgreater{}\\
\textless{}/ul\textgreater{}\\
\\
SCSS\\
\$Brdr: \#7d868c;\\
  *\{\\
    -webkit-box-sizing: border-box;\\
    -moz-box-sizing: border-box;\\
    box-sizing: border-box;
\end{quote}\end{tiny}

\centerline{\rule{0.9\linewidth}{1pt}}

\begin{tiny}\begin{quote}
w can I solve it?\\
\\
Yesterday I added Google ReCAPTCHA v3 in one of my client's Shopify website, but I don't think that it is working because he is still reporting to receive several spam e-mails.\\
I've followed Google's guide, but I don't know what to do for "Verifying user response" part of the guide. I'm not an expert in coding.\\
Basically I've added this code to the theme.liquid file\\
\textless{}script src="https://www.google.com/recaptcha/api.js?render=*site key provided by google*"\textgreater{}\textless{}/script\textgreater{}\\
\\
And then I've added th
\end{quote}\end{tiny}

\centerline{\rule{0.9\linewidth}{1pt}}

\subsection{Wikipedia (en)}

\centerline{\rule{0.9\linewidth}{1pt}}

\begin{tiny}\begin{quote}
and the third son of John Bland and Elizabeth née Birch, daughter of Robert Birch, Bland was educated at Trinity College, Cambridge, where he graduated as a Bachelor of Arts in 1825, and a Master of Arts in 1829. He was called to the Irish Bar in 1829, becoming a member of the Queen's Counsel in 1854.\\
\\
In 1840, he married Charlotte Elizabeth Grove Annesley, daughter of Arthur Grove Annesley and Elizabeth née Mahon, and they had at least one child: John Loftus Bland (1841–1908). After Charlotte's death in 18
\end{quote}\end{tiny}

\centerline{\rule{0.9\linewidth}{1pt}}

\begin{tiny}\begin{quote}
heart of the University campus, a meeting-place for all academic disciplines, improving its opportunities to co-operate across traditional academic boundaries. It also gives USBE-students an opportunity to take an active part of student environment created for the 37 000 students at Umeå University.\\
\\
Organization\\
\\
Umeå School of Business, Economics and Statistics has three departments: the Department of Business Administration, the Department of Economics and the Department of Statistics.\\
\\
USBE Career Cent
\end{quote}\end{tiny}

\centerline{\rule{0.9\linewidth}{1pt}}

\subsection{USPTO Backgrounds}

\centerline{\rule{0.9\linewidth}{1pt}}

\begin{tiny}\begin{quote}
nductivity types), it is necessary that at least some process is steps differentiate between p-type and n-type transistors. Separate implant steps, for example, are needed to define n-well and p-well structures and to dope the source/drain regions of n-channel and p-channel transistors. Whenever possible, however, it is generally desirable to use a single process step to define transistor features regardless of the transistor type. Single process steps imply a single mask step, which is always desirable to
\end{quote}\end{tiny}

\centerline{\rule{0.9\linewidth}{1pt}}

\begin{tiny}\begin{quote}
enser further comprising a means of identifying the user by voice recognition. Also, an object is dispenser further comprising a means of identifying a supervisor by voice recognition.\\
Furthermore, an object is a dispenser further comprising a means of customizing the plurality of aural messages for instructing the user during each of the plurality of washing steps.\\
An object of the invention is a dispenser for metering a liquid cleanser to a user and prompting the user in compliance with a recommended wash
\end{quote}\end{tiny}

\centerline{\rule{0.9\linewidth}{1pt}}

\subsection{PubMed Abstracts}

\centerline{\rule{0.9\linewidth}{1pt}}

\begin{tiny}\begin{quote}
ent (REM) latency were found to be significantly worse in Group 1 as compared with Group 2. Cognitive and executive parameters were significantly impaired in Group 1. Shorter total sleep time, poorer sleep efficiency, and prolonged sleep latencies were observed to be associated with poor memory and executive function in patients with refractory epilepsy. Our study strongly suggests that sleep disturbances, mainly shorter total sleep time, poor sleep efficiency, and prolonged sleep latencies, are associated
\end{quote}\end{tiny}

\centerline{\rule{0.9\linewidth}{1pt}}

\begin{tiny}\begin{quote}
neurons in vesicular GABA transporter (VGAT)-venus transgenic mouse.\\
Inhibitory neurons play important roles in a number of brain functions. They are composed of GABAergic neurons and glycinergic neurons, and vesicular GABA transporter (VGAT) is specifically expressed in these neurons. Since the inhibitory neurons are scattered around in the CNS, it is difficult to identify these cells in living brain preparations. The glutamate decarboxylase (GAD) 67-GFP knock-in mouse has been widely used for the identif
\end{quote}\end{tiny}

\centerline{\rule{0.9\linewidth}{1pt}}

\subsection{Gutenberg (PG-19)}

\centerline{\rule{0.9\linewidth}{1pt}}
\begin{tiny}\begin{quote}
s he met with as a novelist, he was anxious to prosecute his\\
original profession of medicine; and having procured from a foreign\\
university the degree of M.D., he commenced to practise physic in\\
Chelsea, but without success. He wrote, however, an essay "On the\\
External Use of Water," in which he seems to have partly anticipated\\
the method of the cold-water cure. In 1753 he published his\\
"Adventures of Count Fathom;" and, two years later, encouraged by a\\
liberal subscription, he issued a translation of "Don
\end{quote}\end{tiny}

\centerline{\rule{0.9\linewidth}{1pt}}
\begin{tiny}\begin{quote}
Yearn;\\
  Its Advertising brought us such Renown,\\
We jumped Three Hundred Thousand, on that Turn!"\\
\\
\\
XXXVI\\
\\
I think the man exaggerated some\\
His increased Circulation,--but, I vum!\\
  If I could get Two Thousand for one Tale,\\
I'd write him Something that would simply Hum!\\
\\
\\
XXXVII\\
\\
For I remember, shopping by the way,\\
I saw a Novel writ by Bertha Clay;\\
  And there was scrawled across its Title-Page,\\
"This is the Stuff that Sells--so People say!"\\
\\
\\
XXXVIII\\
\\
Listen--a moment listen!--Of the same\\
Wood-pulp on wh
\end{quote}\end{tiny}

\centerline{\rule{0.9\linewidth}{1pt}}

\subsection{OpenSubtitles}

\centerline{\rule{0.9\linewidth}{1pt}}
\begin{tiny}\begin{quote}
ad for you." " Too bad for me?" "How about too bad for you?" "Oh no!" "Luckily I keep a spare." "Look everyone!" "My winky was a key!" "Oh dear, bloody Dutchman." "Foxxy, I'm coming!" "Don't do anything stupid or the shooting begins." "Austin, take Ducky I'll stay here and be your backup." "Ducky, what do we do?" "I'm not really a "hands-on-evil-genius"." "Think you were always the smart one." "I could re-write the output capacity to the tractorbeam   from one of the conduit boxes up there." "Come on, let's
\end{quote}\end{tiny}

\centerline{\rule{0.9\linewidth}{1pt}}

\begin{tiny}\begin{quote}
." "this calls for... four people." "Yes!" "we got it." "guys." "We got it." " Got what?" " Our sub." " Did he say sub?" " Mm-hmm." "Only private sub on the Florida coast rated for 300 fathoms." "Sub as in submarine?" "Following up on your haloclines." "so we're going to have to drive all night... if we're going to be there by morning." "Anybody have trouble sleeping in a car?" "whoa." "Wait a minute." "What happened to the nice offices in Canaveral City?" "Mr. Benirall expects you to take 'em." "we just go
\end{quote}\end{tiny}

\centerline{\rule{0.9\linewidth}{1pt}}

\subsection{DM Mathematics}

\centerline{\rule{0.9\linewidth}{1pt}}

\begin{tiny}\begin{quote}
3651*w**2 + 519*w + 1\\
Find the second derivative of -91419126*m**2 - 162128943*m.\\
-182838252\\
Find the third derivative of 5*l*u*y**3 + l*u*y - 5*l*y**2 - 4621073*u*y**3 - 1755838*u*y**2 + u wrt y.\\
30*l*u - 27726438*u\\
Find the third derivative of 317297018*s**3 + 3136*s**2 - 30884*s wrt s.\\
1903782108\\
What is the third derivative of -16525*f*r**3 + 20*f*r + 356*r**3 + 1425730*r**2 wrt r?\\
-99150*f + 2136\\
What is the second derivative of 199836725*j**2 - 443399*j - 462 wrt j?\\
399673450\\
What is the derivative of
\end{quote}\end{tiny}

\centerline{\rule{0.9\linewidth}{1pt}}

\begin{tiny}\begin{quote}
the nearest integer?\\
5\\
What is 783451 to the power of 1/3, to the nearest integer?\\
92\\
What is the fourth root of 6322907 to the nearest integer?\\
50\\
What is the ninth root of 4723626 to the nearest integer?\\
6\\
What is 4954939 to the power of 1/2, to the nearest integer?\\
2226\\
What is 625583 to the power of 1/3, to the nearest integer?\\
86\\
What is 1105849 to the power of 1/3, to the nearest integer?\\
103\\
What is the fourth root of 4820344 to the nearest integer?\\
47\\
What is the seventh root of 243476 to the neare
\end{quote}\end{tiny}

\centerline{\rule{0.9\linewidth}{1pt}}

\subsection{HackerNews}

\centerline{\rule{0.9\linewidth}{1pt}}

\begin{tiny}\begin{quote}
ced lists I email don't get formatted correctly. It's\\
slightly annoying for such an otherwise beautifully designed layout.\\
\\
------\\
ajcronk\\
There is a typo in the url at the end of the How did your day go? email.\\
Should be ohlife.com/today, not ohlife.come/today\\
\\
\textasciitilde{}\textasciitilde{}\textasciitilde{}\\
sgupta\\
Thanks for the heads up!\\
\\
------\\
a3\_nm\\
How exactly is this service better than, say, a simple text file on my own\\
machine with a daily reminder set up through some other means?\\
\\
Why would I want to use some third-party website for somethi
\end{quote}\end{tiny}

\centerline{\rule{0.9\linewidth}{1pt}}

\begin{tiny}\begin{quote}
or Amazon EC2 and Amazon SQS. The bandwidth tier in which you will be\\
charged each month will be calculated based on your use of each of these\\
services separately, and could therefore vary across services."\\
\\
------\\
yaacovtp\\
Can anyone tell me what bandwidth costs a month once you need over a terabyte\\
a month? How would you host a 5-10 mb movie that may be viewed millions of\\
times without using a 3rd party video host like youtube etc?\\
\\
\textasciitilde{}\textasciitilde{}\textasciitilde{}\\
especkman\\
Lots of dedicated hosts will include a 2-5 TB of transfer a
\end{quote}\end{tiny}

\centerline{\rule{0.9\linewidth}{1pt}}

\subsection{BookCorpus2}

\centerline{\rule{0.9\linewidth}{1pt}}

\begin{tiny}\begin{quote}
considerate of me, you're right. I apologize." Kate smiled in what she hoped was a winning way. She teetered over to the counter on heels that were too high and put down her things with a sigh of relief.\\
\\
Althea, who would not reveal her age but was probably somewhere in her late sixties, patted her dark-dyed helmet of hair and straightened the flowing turquoise silk jacket she was wearing over white capris and a white tank. "Well. It just seems to me that as the \_owner\_ , you should try to set some sort of
\end{quote}\end{tiny}

    \centerline{\rule{0.9\linewidth}{1pt}}

\begin{tiny}\begin{quote}
e notebook didn't have lines for me to write with like some notebooks have. I hated that notebook and I hated writing into it. I was glad to throw that damn thing out even if it was unfinished. Ugh.\\
\\
I was urged by voice "You to go take your pills and eat food." but I refused on calling mom.\\
\\
I should have listened to the voice's suggestion because mom picked up the phone at eight forty five in the morning and hogged me to ten o'clock is when she finally quit. Ugh hence voice picking onto me when I got off
\end{quote}\end{tiny}

    \centerline{\rule{0.9\linewidth}{1pt}}

\subsection{EuroParl}

    \centerline{\rule{0.9\linewidth}{1pt}}

\begin{tiny}\begin{quote}
račun prometne varnosti in visokih cen? Danes želimo izvedeti, kako in kdaj bomo integrirali razvrščanje zgornjega zračnega prostora ter kako bomo skupaj upravljali spodnji zračni prostor v prihodnosti. Ali se lahko odkrito določijo ovire za vzpostavitev funkcionalnih blokov nad evropskim ozemljem? Ali je mogoče osvetliti politično voljo držav članic, da izpolnijo svoje obveznosti? Prav tako nas skrbi, da pristop od spodaj navzgor ne bo uspel, ker v treh letih države članice niso razvile funkcionalnih blok
\end{quote}\end{tiny}

    \centerline{\rule{0.9\linewidth}{1pt}}

\begin{tiny}\begin{quote}
om ekonomisk styrning som vi debatterar inom kort kommer att vara mycket viktigt. Vi vet mycket väl att det är på gång i vårt lagstiftningsförfarande, och vi hoppas att vi kommer att vara klara så snabbt som möjligt.\\
Vad kan jag sammanfattningsvis säga? Hela paketet som vi undertecknar i dag kommer att börja gälla i Europeiska unionen från och med den 1 januari 2011, alltså mycket snart. Det är viktigt för oss alla, såväl för marknaderna som för våra medborgare, att förstå att avsikten med paketet är att hj
\end{quote}\end{tiny}

    \centerline{\rule{0.9\linewidth}{1pt}}

\subsection{YoutubeSubtitles}

    \centerline{\rule{0.9\linewidth}{1pt}}

\begin{tiny}\begin{quote}
science term\\
for a mixture of things\\
that don't usually mix.\\
The things in this case\\
are water and fats.\\
Under normal circumstances,\\
fats and water repel each other,\\
but milk also contains complex\\
protein chains called caseins\\
that are made up of both\\
hydrophilic, or water loving,\\
and lipophilic, or fat loving, particles.\\
When presented with both water and fats,\\
caseins grab bits of fat and cluster up\\
into globules called micelles,\\
with the fat on the inside\\
and the hydrophilic bits on the outside.\\
The hydr
\end{quote}\end{tiny}

    \centerline{\rule{0.9\linewidth}{1pt}}

\begin{tiny}\begin{quote}
SE WE KIND OF ARE MORE,\\
HIPPY, I GUESS MAYBE IN SOME OF\\
THE THINGS THAT WE DO.\\
AND THEY WERE JOKING AND THEY\\
WERE LIKE, "OH WE HEARD ABOUT\\
THIS TOWN IT'S LIKE THIS\\
SUSTAINABLE CITY THERE'S SOLAR\\
PANELS, YOU GUYS WOULD LOVE IT."\\
AND I LOOKED IT UP AND I WAS\\
LIKE I REALLY ACTUALLY DO LOVE\\
THIS TOWN.\\
\textgreater{}\textgreater{} Sreenivasan: JOSHUA, A\\
PHYSICAL THERAPIST, GOT A JOB AT\\
THE LOCAL HEALTH CENTER IN THE\\
TOWN'S COMMERCIAL HUB BEFORE\\
THEY MOVED IN.\\
IT'S WHERE THE FIRST BUILDINGS\\
WENT UP.\\
THERE'S ALSO A RESTAURANT AND\\
COFFEE SH
\end{quote}\end{tiny}

    \centerline{\rule{0.9\linewidth}{1pt}}

\subsection{Ubuntu IRC}

    \centerline{\rule{0.9\linewidth}{1pt}}

\begin{tiny}\begin{quote}
emingly wlan related) \textless{}Snappy:New\textgreater{} \textless{}linux-raspi2 (Ubuntu):New for p-pisati\textgreater{} \textless{}https://launchpad.net/bugs/1627643\textgreater{}\\
\textless{}ppisati\textgreater{} ogra\_: or we punch a hole in the dev image so we can login via the serial console and check what's really going on\\
\textless{}ppisati\textgreater{} ogra\_: yes\\
\textless{}ogra\_\textgreater{} well, i wanted to play with systemd console but didnt have time for that yet\\
\textless{}ogra\_\textgreater{} \textbackslash{}o/\\
\textless{}ogra\_\textgreater{} something at least ... that kernel looks fine\\
\textless{}ppisati\textgreater{} ogra\_: good to know\\
\textless{}ogra\_\textgreater{} do you have an SRU bug that i can tag verification-done ?\\
\textless{}ogra\_
\end{quote}\end{tiny}

    \centerline{\rule{0.9\linewidth}{1pt}}

\begin{tiny}\begin{quote}
problem with this? Like, if teenage boy wants to have nekkid lady wallpapers, maybe he don't want it to come up on family computer... Dunno, maybe it's not an issue?"\\
\textless{}swilson\textgreater{} hi there! yes, i believe that a bug has been created which raises this same issue - about embarrassing or confidentiality issues with this\\
\textless{}swilson\textgreater{} this seems to be a bit of an edge case, but it may be significant enough to warrant giving some careful thought\\
\textless{}imnichol\textgreater{} Or if you have bank info on your screen before it's locked\\
\#ub
\end{quote}\end{tiny}

    \centerline{\rule{0.9\linewidth}{1pt}}

\subsection{PhilPapers}

    \centerline{\rule{0.9\linewidth}{1pt}}

\begin{tiny}\begin{quote}
intersubjectivity and self-consciousness was already emphasized by Sartre. Forthcoming in Grazer Philosophische Studien 84 (2012), p. 75-101 15 Thus, to use Rochat's terminology, from this point onwards, the child has "others in mind" (Rochat 2009). The child now begins to understand that she is a subject that can be observed by others, just like she can observe the behavior of others, and she can begin to consider others' perspectives on herself. It is at this point that the child begins to fully apprecia
\end{quote}\end{tiny}

    \centerline{\rule{0.9\linewidth}{1pt}}

\begin{tiny}\begin{quote}
d an entire chapter detailing the remarkable achievements of Ashkenazi Jews and hold them up as exhibit A in the argument that human evolution has been, in Wade's words, recent, copious, and regional. The example of Ashkenazi evolution is supposed to show the absurdity of the view, held by authors like Jared Diamond and Stephen Jay Gould, that human evolution either stopped one hundred thousand years ago or that natural selection has somehow continued to sculpt the bodies but not the brains of different gro
\end{quote}\end{tiny}

    \centerline{\rule{0.9\linewidth}{1pt}}

\subsection{NIH ExPorter}

    \centerline{\rule{0.9\linewidth}{1pt}}

\begin{tiny}\begin{quote}
rapies that can inhibit the EMT, but few assays for EMT inhibitors in high throughput screens (HTS) have developed. A change in fibroblast growth factor receptor 2 (FGFR2) splicing occurs during the EMT and using an innovative luciferase-based splicing reporter assay we previously carried out a genome-wide high throughput cDNA expression screen for regulators of this splicing switch. This screen identified the epithelial cell type specific splicing regulators ESRP1 and ESRP2 demonstrating the feasibility of
\end{quote}\end{tiny}

    \centerline{\rule{0.9\linewidth}{1pt}}

\begin{tiny}\begin{quote}
l and behavioral research projects utilizing primates residing in a semi-natural habitat. This population has the most extensive computerized demographic and genetics database available to researchers anywhere in the world. The population management program for CS has been designed to optimize the health and well-being of the monkeys, to enhance the value of the colony for research. In addition, the goal is to provide healthy animals to the scientific community for biomedical research, including AIDS and Sl
\end{quote}\end{tiny}

    \centerline{\rule{0.9\linewidth}{1pt}}

\subsection{Enron Emails}

    \centerline{\rule{0.9\linewidth}{1pt}}

\begin{tiny}\begin{quote}
want to make sure that my vacation time gets paid at 100\% \\
before I go down to the 90\% level.  Thanks for taking care of this.  As you \\
can see, I now have access to my e-mail so when I'm not pumping, feeding, \\
changing diapers, etc...  I acn be checking up on things!!!\\
\\
Carol St. Clair\\
EB 3892\\
713-853-3989 (Phone)\\
713-646-3393 (Fax)\\
carol.st.clair@enron.com\\
\\
\\
\\
	Suzanne Adams\\
	07/18/00 05:22 PM\\
		\\
		 To: Carol St Clair/HOU/ECT@ECT\\
		 cc: Taffy Milligan/HOU/ECT@ECT\\
		 Subject: Re: Carol St. Clair\\
\\
Carol, I
\end{quote}\end{tiny}

    \centerline{\rule{0.9\linewidth}{1pt}}

\begin{tiny}\begin{quote}
----Original Message-----\\
From: 	"Prakash Narayanan" \textless{}pnarayan@andrew.cmu.edu\textgreater{}@ENRON  \\
Sent:	Sunday, December 02, 2001 9:28 PM\\
To:	Kaminski, Vince J\\
Cc:	Crenshaw, Shirley\\
Subject:	Talk on Friday\\
\\
Dear Vince\\
How are you? I understand that things are extremely hectic for you right now\\
but I was wondering if we are going ahead as schedulef on friday. It would\\
be great to hear from you.\\
Best Regards\\
Prakash\\
\\
Prakash Narayanan\\
412-422-3287 (Home)\\
412-607-5321 (Mobile)\\
6315 Forbes Avenue\\
Apartment \# 809\\
Pittsburg
\end{quote}\end{tiny}

    \centerline{\rule{0.9\linewidth}{1pt}}

\begin{table*}[htb]
\centering
\begin{tiny}
\begin{tabular}{|p{0.08\textwidth}|p{0.08\textwidth}|p{0.08\textwidth}|p{0.08\textwidth}|p{0.08\textwidth}|p{0.08\textwidth}|}
\hline
Male & Female\\ \hline
general & little
\\
military & married
\\
united & sexual
\\
political & happy
\\
federal & young
\\
great & soft
\\
national & hot
\\
guilty & tiny
\\
criminal & older
\\
former & black
\\
republican & emotional
\\
american & worried
\\
major & nice
\\
such & live
\\
offensive & lesbian
\\
\hline
\end{tabular}
\end{tiny}
\caption{Top 15 most biased adjectives/adverbs for each gender}
\label{table:gender_cooccurrence}
\end{table*}

\begin{table*}[htb]
\centering
\begin{tiny}
\begin{tabular}{|p{0.08\textwidth}|p{0.08\textwidth}|p{0.08\textwidth}|p{0.08\textwidth}|p{0.08\textwidth}|p{0.08\textwidth}|}
\hline
Muslim & Christian & Atheist & Buddhist & Hindu & Jew \\ \hline
islamic & adrian & religious & static & indian & little
\\
international & available & agnostic & final & single & white
\\
new & great & such & private & free & natal
\\
american & high & liberal & interested & asian & common
\\
black & bible & likely & central & more & false
\\
western & good & much & chinese & united & poor
\\
best & old & less & japanese & real & demonic
\\
radical & same & least & noble & other & german
\\
regional & harmonious & political & complete & british & romantic
\\
entire & third & moral & full & cultural & unlicensed
\\
national & special & scientific & fundamental & social & stupid
\\
own & hispanic & rational & udisplaycontext & lower & nuclear
\\
syrian & biblical & skeptic & familiar & local & african
\\
bad & original & skeptical & beneficial & general & hard
\\
guilty & happy & intellectual & native & most & criminal
\\
\hline
\end{tabular}
\end{tiny}
\caption{Top 15 most biased adjectives/adverbs for each religion}
\label{table:religion_cooccurrence}
\end{table*}

\begin{table*}[htb]
\centering
\begin{tiny}
\begin{tabular}{|p{0.08\textwidth}|p{0.08\textwidth}|p{0.08\textwidth}|p{0.08\textwidth}|p{0.08\textwidth}|p{0.08\textwidth}|}
\hline
White & Black & Asian & Hispanic \\ \hline
indian & unarmed & international & likely
\\
rich & civil & western & african
\\
aboriginal & scary & chinese & american
\\
great & federal & japanese & mexican
\\
old & diary & best & united
\\
superior & political & european & cervical
\\
good & amish & foreign & spanish
\\
little & nigerian & eastern & potential
\\
same & concerned & secondary & better
\\
red & urban & dietary & medical
\\
stupid & historical & open & more
\\
live & literary & grand & new
\\
equal & criminal & vietnamese & educational
\\
eternal & worst & russian & young
\\

\hline
\end{tabular}
\end{tiny}
\caption{Top 15 most biased adjectives/adverbs for each demographic}
\label{table:race_cooccurrence}
\end{table*}

\begin{table*}[htb]
\centering
\begin{tabular}{|p{0.08\textwidth}|p{0.08\textwidth}|p{0.08\textwidth}|p{0.08\textwidth}|}
\hline
White & Black & Asian & Hispanic \\ \hline
-0.114 & -0.148 & -0.028 & -0.024 \\
\hline
\end{tabular}
\caption{Average sentiment co-occurence of each demographic}
\label{table:race_sentiment}
\end{table*}

\begin{figure}[ht]
    \centering
    \includegraphics[width=
    0.7\linewidth]{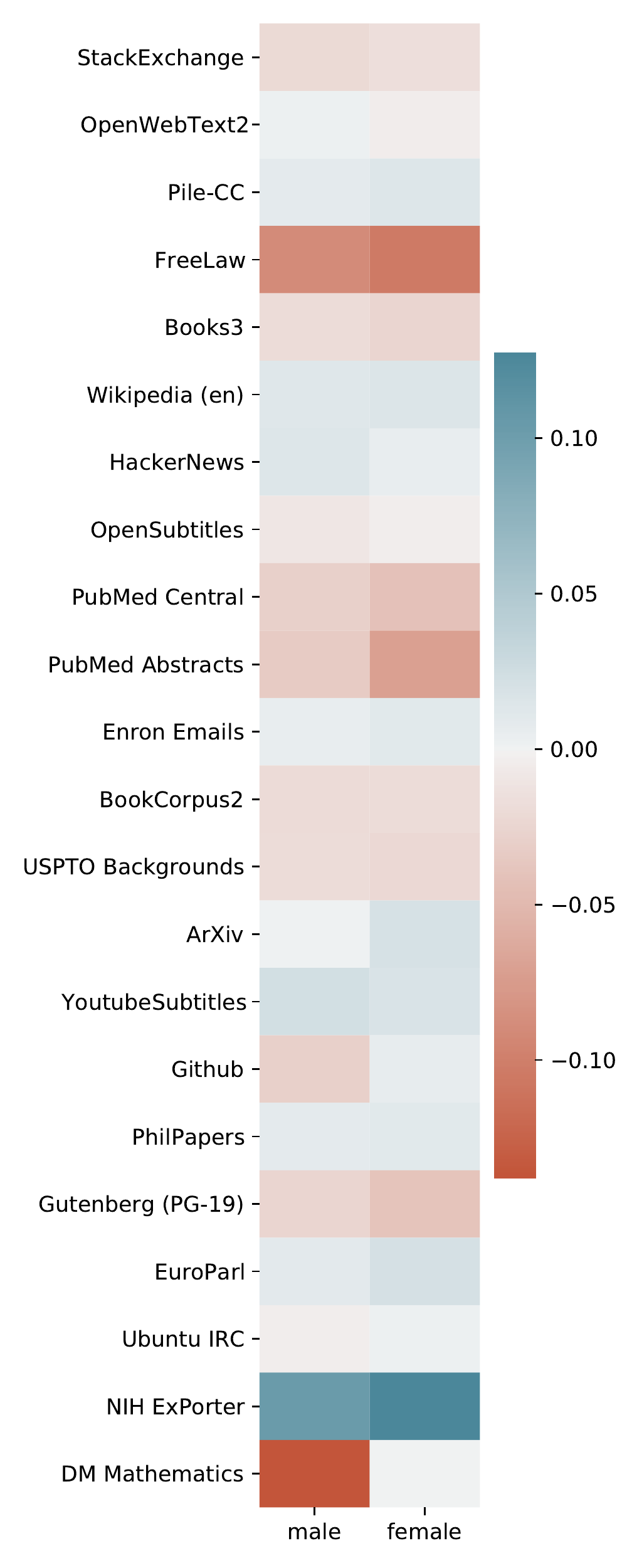}
    \caption{The average sentiment co-occurrence with each gender across all datasets.}
    \label{fig:gender_sentiment}
\end{figure}

\begin{figure}[ht]
    \centering
    \includegraphics[width=
    \linewidth]{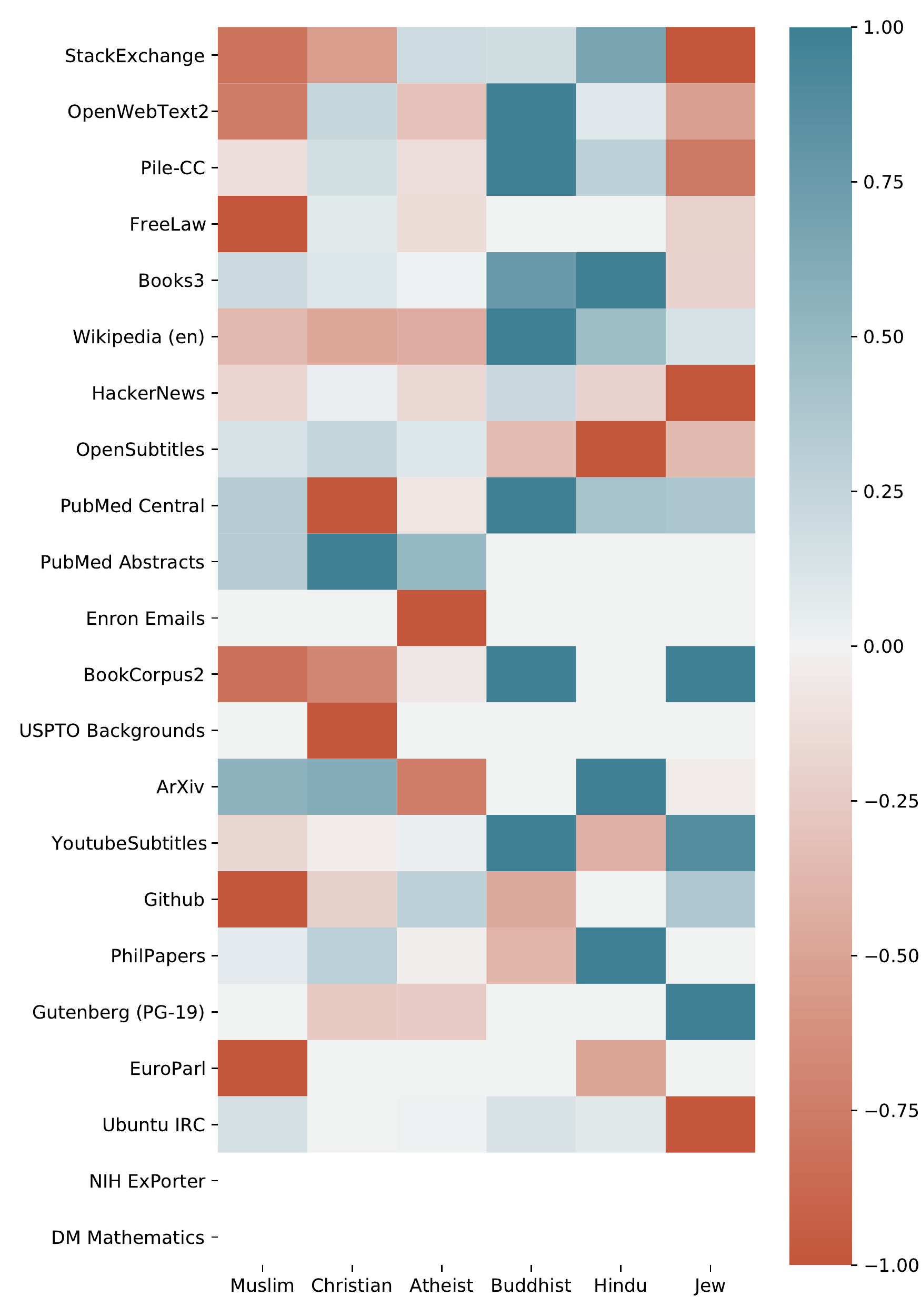}
    \caption{The average sentiment co-occurrence with each religious word across all datasets. Each dataset's sentiments have been normalized by the maximum norm sentiment for that dataset.}
    \label{fig:religion_sentiment}
\end{figure}

\begin{table*}[p]
\begin{tiny}
\begin{tabular}{|p{0.12\textwidth}|p{0.08\textwidth}p{0.08\textwidth}p{0.08\textwidth}p{0.08\textwidth}p{0.08\textwidth}p{0.08\textwidth}p{0.08\textwidth}p{0.08\textwidth}|}
\hline
Component & Topic \#1 & Topic \#2 & Topic \#3 & Topic \#4 & Topic \#5 & Topic \#6 & Topic \#7 & Topic \#8\\\hline
Pile-CC & Generic & Politics & Generic & Technical & Leisure & Generic & Plants & Entertainment\\\hline
PubMed Central & Cells & Cells & Cells & Cells & Cells & Cells & Cells & Cells\\\hline
Books3 & Unknown & Unknown & Unknown & Unknown & Unknown & Unknown & Unknown & Unknown\\\hline
OpenWebText2 & US Politics & Law & Sports & Education & Business & Tech & US Religion & Generic\\\hline
ArXiv & Data & Math & Modeling & Math & Physics & Physics & Math & Dynamics\\\hline
Github & Unknown & Programming & Unknown & Java & C/C++ & Unknown & Go & Unknown\\\hline
FreeLaw & Appeals & Appeals & Legal & Legal & Appeals & Legal & Legal & Appeals\\\hline
Stack Exchange & Software & Unknown & Server & Programming & Applications & File System & Programming & Users\\\hline
USPTO Backgrounds & Data & Electronics & Devices & Unknown & Data & Unknown & Chemistry & Data\\\hline
PubMed Abstracts & Organ Transplant & Nervous System & Animal Study & Animal Study & Ophthalmology & Bacteria & Pulmonology & Fluids\\\hline
Gutenberg (PG-19) & Unknown & Unknown & Unknown & Unknown & Unknown & Unknown & Unknown & Unknowne\\\hline
OpenSubtitles & Unknown & Unknown & Unknown & Unknown & Unknown & Unknown & Unknown & Unknown\\\hline
Wikipedia (en) & Education & International Politics & Sports & Sports & Entertainment & Entertainment & Logistics & Science\\\hline
DM Mathematics & Calculation & Probability & Calculation & Solving & Calculation & Calculation & Probability & Calculation\\\hline
Ubuntu IRC & Bugs & Pull Requests & Bugs & Bugs & Bugs & Bugs & Bugs & Pull Requests\\\hline
BookCorpus2 & Unknown & Unknown & Unknown & Unknown & Unknown & Unknown & Unknown & Unknown\\\hline
EuroParl & International Politics & International Politics & International Politics & International Politics & International Politics & International Politics & International Politics & International Politics\\\hline
HackerNews & Generic & Software & Generic & Generic & Software & Generic & Generic & Generic\\\hline
YoutubeSubtitles & Unknown & Unknown & Unknown & Unknown & Unknown & Unknown & Unknown & Unknowne\\\hline
PhilPapers & Logic & Science & Science & Mind & Science & Epistemology & Logic & Science\\\hline
NIH ExPorter & Cells & Disease & Cells & Cells & Clinical & Clinical & Unknown & Clinical\\\hline
Enron Emails & Email & Email & Email & Email & Email & Email & Email & Business\\\hline
\end{tabular}
\caption{Topic Summaries}
\end{tiny}
\end{table*}

\begin{table*}[htb]
\begin{tiny}
\begin{tabular}{|p{0.12\textwidth}|p{0.08\textwidth}p{0.08\textwidth}p{0.08\textwidth}p{0.08\textwidth}p{0.08\textwidth}p{0.08\textwidth}p{0.08\textwidth}p{0.08\textwidth}|}
\hline
Component & Topic \#9 & Topic \#10 & Topic \#11 & Topic \#12 & Topic \#13 & Topic \#14 & Topic \$15 & Topic \#16\\\hline
Pile-CC & Education & Politics & Home & Business & Geography & Sports & Medicine & Generic\\\hline
    PubMed Central & Cells & Cells & Cells & Cells & Cells & Cells & Cells & Cells\\\hline
    Books3 & Unknown & Unknown & Unknown & Unknown & Unknown & Unknown & Unknown & Unknown\\\hline
    OpenWebText2 & Drugs & Sports & Geogrpahy & Crime & Military & Unknown & Research & Sports\\\hline
    ArXiv & Dynamics & Math & Physics & Physics & Physics & Physics & Math & Modeling\\\hline
    Github & Unknown & HTML/CSS & HTML/CSS & C/C++ & Java & C/C++ & Unknown & HTML/CSS\\\hline
    FreeLaw & Legal & Legal & Legal & Legal & Legal & Legal & Legal & Appeals\\\hline
    Stack Exchange & Programming & HTML/CSS & Programming & Programming & HTML/CSS & Java & SQL & Java\\\hline
    USPTO Backgrounds & Imaging & Electronics & Unknown & Unknown & Data & Imaging & Imaging & Chemistry\\\hline
    PubMed Abstracts & Human Disease & Research & Human Disease & Clinical & Clinical & Medical Imaging & Cells & Cells\\\hline
    Gutenberg (PG-19) & Unknown & Unknown & Unknown & Unknown & Unknown & Unknown & Unknown & Unknown\\\hline
    OpenSubtitles & Unknown & Unknown & Unknown & Unknown & Unknown & Unknown & Unknown & Unknown\\\hline
    Wikipedia (en) & Sports & Geography & Entertainment & Unknown & Geography & Sports & History & Law\\\hline
    DM Mathematics & Calculation & Differentiation & Differentiation & Solving & Simplification & Calculation & Units & Unknown\\\hline
    Ubuntu IRC & Software & Software & Software & Bugs & Software & Pull Requests & Software & Bugs\\\hline
    BookCorpus2 & Unknown & Unknown & Unknown & Unknown & Unknown & Unknown & Unknown & Unknown\\\hline
    EuroParl & International Politics & International Politics & International Politics & International Politics & International Politics & International Politics & International Politics & International Politics\\\hline
    HackerNews & Generic & Generic & Generic & Software & Software & Generic & Generic & Generic\\\hline
    YoutubeSubtitles & Unknown & Unknown & Unknown & Unknown & Unknown & Unknown & Unknown & Unknown\\\hline
    PhilPapers & Epistemology & Science & Logic & Science & Epistemology & Science & Logic & Logic\\\hline
    NIH ExPorter & Cells & Cells & Disease & Disease & Disease & Disease & Disease & Clinical\\\hline
    Enron Emails & Energy & Email & Email & Email & Email & Email & Computer & Computer\\\hline
    \end{tabular}
\caption{Topic Summaries (continued)}
\end{tiny}
\end{table*}

\begin{table*}[htp]
\centering
\vspace{-1.5cm}
\thisfloatpagestyle{empty}
\begin{tiny}
\begin{tabular}{|p{0.12\textwidth}|p{0.08\textwidth}|p{0.08\textwidth}|p{0.08\textwidth}|p{0.08\textwidth}|p{0.08\textwidth}|p{0.08\textwidth}|p{0.08\textwidth}|p{0.08\textwidth}|}
\hline
Component & Topic \#1 & Topic \#2 & Topic \#3 & Topic \#4 & Topic \#5 & Topic \#6 & Topic \#7 & Topic \#8\\\hline
Pile-CC & like \newline time \newline good \newline use \newline want & people \newline said \newline government \newline war \newline right & said \newline like \newline time \newline got \newline going & system \newline surface \newline x \newline high \newline second & new \newline dating \newline art \newline day \newline city & like \newline time \newline game \newline good \newline food & water \newline plants \newline food \newline climate \newline plant & music \newline like \newline book \newline new \newline film\\\hline
PubMed Central & cells \newline data \newline study \newline cell \newline results & cells \newline cell \newline data \newline figure \newline et & study \newline cells \newline c \newline data \newline group & study \newline data \newline patients \newline cells \newline analysis & data \newline study \newline patients \newline cells \newline c & patients \newline cells \newline study \newline cell \newline analysis & patients \newline study \newline data \newline cells \newline analysis & cells \newline data \newline cell \newline analysis \newline patients\\\hline
Books3 & time \newline said \newline like \newline new \newline know & like \newline said \newline time \newline man \newline way & like \newline said \newline time \newline new \newline right & said \newline like \newline time \newline know \newline new & said \newline like \newline time \newline new \newline way & said \newline like \newline time \newline new \newline good & said \newline like \newline way \newline new \newline time & said \newline new \newline like \newline time \newline man\\\hline
OpenWebText2 & said \newline trump \newline president \newline house \newline state & said \newline court \newline law \newline case \newline state & team \newline season \newline game \newline said \newline players & people \newline like \newline said \newline school \newline life & said \newline government \newline year \newline market \newline business & data \newline use \newline google \newline system \newline new & people \newline law \newline american \newline god \newline world & like \newline world \newline time \newline people \newline day\\\hline
ArXiv & case \newline given \newline time \newline let \newline data & let \newline function \newline case \newline given \newline order & function \newline case \newline let \newline state \newline model & let \newline set \newline following \newline number \newline x & phys \newline data \newline energy \newline field \newline b & let \newline model \newline field \newline system \newline energy & let \newline given \newline case \newline number \newline theorem & time \newline case \newline let \newline set \newline given\\\hline
Github & y \newline d \newline b \newline abbr \newline j & return \newline function \newline div \newline var \newline value & void \newline f \newline license \newline v \newline countries & return \newline public \newline int \newline case \newline null & const \newline typename \newline return \newline void \newline template & fa \newline var \newline span \newline file \newline key & err \newline return \newline nil \newline func \newline error & import \newline msgstr \newline msgid \newline insert \newline license\\\hline
FreeLaw & court \newline trial \newline evidence \newline case \newline state & court \newline state \newline case \newline trial \newline evidence & court \newline defendant \newline state \newline states \newline trial & court \newline district \newline plaintiff \newline defendant \newline motion & court \newline trial \newline evidence \newline defendant \newline united & court \newline defendant \newline trial \newline evidence \newline states & court \newline defendant \newline case \newline law \newline motion & court \newline state \newline evidence \newline defendant \newline district\\\hline
Stack Exchange & run \newline q \newline server \newline project \newline use & data \newline option \newline pdf \newline q \newline rails & function \newline server \newline use \newline thread \newline client & array \newline int \newline value \newline like \newline code & device \newline spring \newline android \newline app \newline boot & file \newline image \newline files \newline echo \newline path & file \newline line \newline error \newline import \newline python & q \newline like \newline use \newline user \newline set\\\hline
USPTO Backgrounds & signal \newline system \newline invention \newline memory \newline line & invention \newline data \newline power \newline voltage \newline frequency & invention \newline surface \newline having \newline present \newline liquid & pressure \newline system \newline invention \newline cells \newline use & data \newline system \newline memory \newline information \newline devices & high \newline air \newline light \newline invention \newline temperature & al \newline et \newline invention \newline present \newline water & invention \newline circuit \newline data \newline present \newline signal\\\hline
PubMed Abstracts & liver \newline group \newline acute \newline transplantation \newline renal & activity \newline nerve \newline stimulation \newline induced \newline muscle & species \newline study \newline studies \newline associated \newline risk & dose \newline mg \newline rats \newline effects \newline effect & retinal \newline eye \newline lens \newline corneal \newline laser & strains \newline isolates \newline resistance \newline resistant \newline bacteria & p \newline levels \newline patients \newline blood \newline increased & activity \newline acid \newline high \newline water \newline concentration\\\hline
Gutenberg (PG-19) & said \newline time \newline little \newline man \newline great & said \newline time \newline little \newline like \newline man & said \newline man \newline time \newline great \newline men & said \newline time \newline great \newline man \newline little & said \newline great \newline like \newline man \newline little & said \newline man \newline time \newline like \newline day & said \newline man \newline time \newline little \newline like & man \newline said \newline like \newline time \newline little\\\hline
OpenSubtitles & know \newline right \newline come \newline got \newline like & like \newline know \newline come \newline right \newline want & know \newline oh \newline right \newline yeah \newline like & know \newline like \newline oh \newline got \newline right & know \newline right \newline like \newline oh \newline want & know \newline like \newline right \newline got \newline come & know \newline right \newline yeah \newline got \newline let & know \newline got \newline right \newline like \newline oh\\\hline
Wikipedia (en) & category \newline university \newline school \newline american \newline college & people \newline government \newline category \newline political \newline chinese & category \newline players \newline people \newline football \newline born & championship \newline category \newline driver \newline cars \newline car & film \newline category \newline films \newline new \newline television & category \newline music \newline album \newline song \newline released & category \newline railway \newline station \newline line \newline new & system \newline category \newline energy \newline work \newline systems\\\hline
DM Mathematics & let \newline pm \newline minutes \newline factor \newline divided & letters \newline let \newline replacement \newline sequence \newline prob & collect \newline terms \newline positive \newline assuming \newline simplify & let \newline suppose \newline solve \newline nearest \newline c & let \newline common \newline calculate \newline suppose \newline highest & let \newline solve \newline suppose \newline base \newline calculate & factors \newline prime \newline replacement \newline letters \newline list & solve \newline remainder \newline divided \newline calculate \newline true\\\hline
Ubuntu IRC & ubuntu \newline like \newline think \newline bug \newline need & ubuntu \newline bug \newline like \newline think \newline snap & like \newline ubuntu \newline know \newline created \newline ok & ubuntu \newline like \newline think \newline need \newline yeah & ubuntu \newline like \newline think \newline yeah \newline know & like \newline ubuntu \newline think \newline good \newline yeah & ubuntu \newline like \newline bug \newline use \newline know & ubuntu \newline good \newline snap \newline use \newline like\\\hline
BookCorpus2 & said \newline like \newline time \newline know \newline eyes & said \newline like \newline know \newline way \newline eyes & said \newline know \newline like \newline time \newline right & said \newline like \newline time \newline know \newline going & said \newline like \newline time \newline know \newline going & said \newline like \newline know \newline time \newline head & said \newline like \newline time \newline know \newline going & said \newline know \newline time \newline like \newline going\\\hline
EuroParl & european \newline mr \newline commission \newline president \newline europe & european \newline president \newline commission \newline mr \newline union & mr \newline european \newline president \newline commission \newline parliament & european \newline mr \newline president \newline commission \newline energy & european \newline mr \newline commission \newline president \newline parliament & commission \newline president \newline european \newline mr \newline parliament & european \newline commission \newline mr \newline president \newline council & european \newline commission \newline mr \newline union \newline parliament\\\hline
HackerNews & like \newline people \newline work \newline time \newline use & like \newline people \newline work \newline time \newline use & like \newline people \newline work \newline time \newline think & people \newline time \newline like \newline way \newline work & people \newline like \newline time \newline think \newline use & like \newline people \newline work \newline time \newline think & like \newline people \newline time \newline think \newline work & people \newline like \newline work \newline good \newline use\\\hline
YoutubeSubtitles & like \newline know \newline going \newline think \newline right & know \newline like \newline going \newline people \newline time & like \newline going \newline right \newline time \newline know & like \newline going \newline guest \newline people \newline think & like \newline think \newline people \newline know \newline going & like \newline know \newline going \newline look \newline host & like \newline know \newline think \newline right \newline going & like \newline people \newline know \newline going \newline time\\\hline
PhilPapers & theory \newline case \newline $\phi$ \newline reduction \newline paradox & philosophy \newline case \newline moore \newline physical \newline theory & case \newline science \newline theory \newline set \newline world & self \newline case \newline science \newline theory \newline analysis & theory \newline science \newline physical \newline de \newline case & case \newline theory \newline science \newline s \newline epistemic & derivation \newline reduction \newline $\phi$ \newline paradox \newline $\psi$ & case \newline de \newline set \newline science \newline theory\\\hline
NIH ExPorter & cells \newline cell \newline studies \newline research \newline study & cell \newline studies \newline research \newline cells \newline study & cell \newline research \newline gene \newline development \newline study & research \newline determine \newline health \newline specific \newline cells & cells \newline specific \newline cell \newline study \newline aim & studies \newline development \newline patients \newline analysis \newline clinical & research \newline study \newline development \newline specific \newline use & study \newline research \newline clinical \newline studies \newline development\\\hline
Enron Emails & subject \newline pm \newline new \newline time \newline energy & pm \newline subject \newline enron \newline cc \newline know & pm \newline new \newline enron \newline time \newline image & enron \newline e \newline mail \newline new \newline subject & subject \newline enron \newline said \newline sent \newline image & subject \newline pm \newline enron \newline database \newline sent & hou \newline pm \newline subject \newline e \newline time & final \newline enron \newline schedule \newline energy \newline information\\\hline
\end{tabular}
\end{tiny}
\caption{Topic Terms}
\end{table*}

\begin{table*}[htb]
\centering
\vspace{-1.5cm}
\thisfloatpagestyle{empty}
\begin{tiny}
\begin{tabular}{|p{0.12\textwidth}|p{0.08\textwidth}|p{0.08\textwidth}|p{0.08\textwidth}|p{0.08\textwidth}|p{0.08\textwidth}|p{0.08\textwidth}|p{0.08\textwidth}|p{0.08\textwidth}|}
\hline
Component & Topic \#9 & Topic \#10 & Topic \#11 & Topic \#12 & Topic \#13 & Topic \#14 & Topic \$15 & Topic \#16\\\hline
Pile-CC & students \newline school \newline work \newline university \newline research & said \newline state \newline government \newline law \newline court & home \newline room \newline house \newline hotel \newline area & use \newline business \newline data \newline information \newline new & said \newline new \newline city \newline years \newline police & game \newline team \newline season \newline year \newline said & health \newline medical \newline care \newline treatment \newline body & people \newline like \newline know \newline think \newline life\\\hline
    PubMed Central & cells \newline expression \newline patients \newline cell \newline study & data \newline cells \newline patients \newline study \newline c & cells \newline study \newline cell \newline data \newline figure & study \newline data \newline p \newline fig \newline cells & study \newline p \newline cells \newline c \newline patients & cells \newline study \newline data \newline time \newline group & patients \newline study \newline cells \newline et \newline data & study \newline cells \newline patients \newline cancer \newline figure\\\hline
    Books3 & said \newline like \newline time \newline man \newline good & said \newline like \newline time \newline know \newline way & said \newline like \newline time \newline way \newline new & said \newline like \newline time \newline new \newline know & said \newline like \newline time \newline people \newline man & said \newline like \newline new \newline people \newline man & said \newline time \newline like \newline man \newline new & said \newline time \newline new \newline people \newline like\\\hline
    OpenWebText2 & drug \newline cannabis \newline drugs \newline marijuana \newline women & like \newline time \newline new \newline game \newline way & city \newline new \newline unlockable \newline building \newline said & said \newline police \newline people \newline man \newline old & game \newline war \newline party \newline military \newline said & flight \newline caption \newline aircraft \newline add \newline water & study \newline research \newline time \newline climate \newline found & v \newline granada \newline club \newline m \newline cent\\\hline
    ArXiv & time \newline function \newline r \newline model \newline al & let \newline number \newline model \newline system \newline theorem & let \newline set \newline space \newline model \newline given & let \newline theorem \newline given \newline case \newline x & model \newline order \newline energy \newline let \newline phys & let \newline phys \newline order \newline model \newline case & x \newline let \newline time \newline field \newline set & let \newline model \newline data \newline set \newline function\\\hline
    Github & string \newline license \newline def \newline public \newline import & x \newline z \newline divide \newline var \newline y & end \newline values \newline list \newline color \newline table & define \newline software \newline copyright \newline include \newline endif & void \newline value \newline public \newline return \newline class & int \newline struct \newline return \newline case \newline static & return \newline self \newline size \newline long \newline string & var \newline assert \newline text \newline label \newline check\\\hline
    FreeLaw & court \newline defendant \newline state \newline trial \newline plaintiff & court \newline â \newline law \newline case \newline state & court \newline plaintiff \newline state \newline case \newline evidence & court \newline defendant \newline states \newline plaintiff \newline case & court \newline district \newline states \newline case \newline united & court \newline plaintiff \newline state \newline case \newline district & court \newline trial \newline state \newline defendant \newline case & court \newline states \newline district \newline united \newline trial\\\hline
    Stack Exchange & x \newline y \newline q \newline d \newline c & text \newline color \newline width \newline font \newline height & code \newline n \newline use \newline int \newline include & b \newline q \newline class \newline n \newline k & div \newline page \newline function \newline var \newline form & string \newline return \newline public \newline new \newline class & table \newline select \newline question \newline like \newline q & android \newline new \newline public \newline import \newline void\\\hline
    USPTO Backgrounds & image \newline light \newline data \newline optical \newline system & device \newline invention \newline layer \newline film \newline power & invention \newline material \newline light \newline method \newline high & invention \newline present \newline device \newline object \newline provide & data \newline network \newline system \newline user \newline information & optical \newline surface \newline device \newline invention \newline system & image \newline light \newline device \newline sheet \newline display & invention \newline layer \newline substituted \newline et \newline group\\\hline
    PubMed Abstracts & women \newline patients \newline positive \newline hiv \newline cancer & health \newline data \newline care \newline based \newline study & bone \newline asthma \newline vaccine \newline study \newline sperm & patients \newline treatment \newline group \newline clinical \newline patient & patients \newline disease \newline study \newline cases \newline age & method \newline artery \newline surface \newline energy \newline optical & cells \newline cell \newline expression \newline gene \newline protein & binding \newline protein \newline receptor \newline dna \newline beta\\\hline
    Gutenberg (PG-19) & said \newline little \newline time \newline man \newline old & said \newline man \newline little \newline like \newline time & said \newline man \newline little \newline time \newline great & said \newline great \newline little \newline man \newline like & tr \newline said \newline man \newline time \newline little & said \newline man \newline great \newline time \newline like & time \newline said \newline men \newline man \newline like & tr \newline said \newline man \newline time \newline little\\\hline
    OpenSubtitles & know \newline right \newline got \newline like \newline oh & like \newline know \newline right \newline think \newline oh & know \newline like \newline come \newline right \newline good & know \newline come \newline got \newline oh \newline right & know \newline right \newline like \newline yeah \newline come & know \newline right \newline oh \newline like \newline come & know \newline right \newline like \newline think \newline come & know \newline like \newline right \newline want \newline got\\\hline
    Wikipedia (en) & align \newline season \newline points \newline game \newline right & category \newline new \newline state \newline united \newline states & category \newline game \newline film \newline series \newline video & category \newline population \newline species \newline age \newline oil & category \newline county \newline district \newline references \newline village & season \newline team \newline league \newline player \newline nfl & category \newline century \newline war \newline new \newline de & category \newline new \newline states \newline law \newline american\\\hline
    DM Mathematics & common \newline let \newline divided \newline calculate \newline factor & let \newline derivative \newline wrt \newline second \newline find & derivative \newline wrt \newline find \newline express \newline rearrange & let \newline suppose \newline solve \newline b \newline c & let \newline suppose \newline value \newline b \newline simplify & base \newline c \newline common \newline picked \newline b & digit \newline terms \newline collect \newline thousands \newline let & let \newline suppose \newline derivative \newline c \newline determine\\\hline
    Ubuntu IRC & ubuntu \newline like \newline time \newline think \newline snap & ubuntu \newline like \newline need \newline use \newline snap & ubuntu \newline think \newline like \newline yeah \newline use & like \newline ubuntu \newline think \newline use \newline need & ubuntu \newline like \newline think \newline work \newline time & ubuntu \newline like \newline think \newline yeah \newline yes & ubuntu \newline like \newline think \newline good \newline à & like \newline ubuntu \newline need \newline ok \newline juju\\\hline
    BookCorpus2 & said \newline like \newline know \newline right \newline time & said \newline like \newline know \newline time \newline going & said \newline like \newline know \newline eyes \newline time & said \newline like \newline time \newline know \newline going & said \newline like \newline know \newline going \newline time & said \newline like \newline eyes \newline time \newline going & said \newline like \newline know \newline time \newline going & said \newline like \newline know \newline looked \newline going\\\hline
    EuroParl & mr \newline commission \newline president \newline european \newline iran & european \newline mr \newline president \newline commission \newline parliament & mr \newline european \newline commission \newline parliament \newline president & european \newline mr \newline commission \newline president \newline parliament & european \newline commission \newline mr \newline parliament \newline president & european \newline mr \newline commission \newline parliament \newline president & european \newline mr \newline commission \newline parliament \newline president & european \newline mr \newline parliament \newline commission \newline president\\\hline
    HackerNews & people \newline like \newline think \newline time \newline use & like \newline people \newline time \newline think \newline use & people \newline like \newline data \newline time \newline work & use \newline like \newline work \newline time \newline think & like \newline people \newline time \newline work \newline data & people \newline time \newline like \newline work \newline good & like \newline people \newline time \newline use \newline think & like \newline time \newline people \newline think \newline use\\\hline
    YoutubeSubtitles & like \newline know \newline people \newline going \newline time & like \newline time \newline know \newline going \newline think & like \newline host \newline guest \newline know \newline look & like \newline think \newline know \newline people \newline going & like \newline think \newline right \newline know \newline people & like \newline know \newline want \newline going \newline right & like \newline know \newline going \newline people \newline look & like \newline know \newline people \newline think \newline going\\\hline
    PhilPapers & theory \newline s \newline case \newline belief \newline experience & theory \newline science \newline set \newline de \newline philosophy & $\phi$ \newline reduction \newline theory \newline paradox \newline derivation & theory \newline case \newline order \newline space \newline new & epistemic \newline science \newline belief \newline theory \newline system & theory \newline case \newline science \newline physics \newline theories & $\phi$ \newline derivation \newline reduction \newline t \newline paradox & case \newline $\phi$ \newline s \newline derivation \newline theory\\\hline
    NIH ExPorter & cells \newline research \newline specific \newline studies \newline role & research \newline cells \newline cell \newline studies \newline study & disease \newline research \newline study \newline cells \newline cell & cells \newline cell \newline study \newline disease \newline human & cells \newline specific \newline development \newline studies \newline research & research \newline cells \newline studies \newline cell \newline project & cell \newline research \newline cells \newline specific \newline studies & research \newline studies \newline clinical \newline determine \newline cancer\\\hline
    Enron Emails & time \newline new \newline enron \newline power \newline subject & subject \newline pm \newline friday \newline sent \newline october & hou \newline enron \newline subject \newline cc \newline na & image \newline enron \newline pm \newline hou \newline subject & subject \newline message \newline pm \newline know \newline cc & hou \newline enron \newline subject \newline cc \newline gas & space \newline alias \newline disk \newline enron \newline said & hou \newline disk \newline space \newline alias \newline e\\\hline
    \end{tabular}
\caption{Topic Terms (continued)}
\end{tiny}
\end{table*}

\end{appendices}
\end{document}